\documentclass[10pt,twocolumn,letterpaper]{article}

\usepackage{iccv}
\usepackage{times}
\usepackage{epsfig}
\usepackage{graphicx}
\usepackage{amsmath}
\usepackage{amssymb}

\usepackage{booktabs}
\usepackage{kotex}
\usepackage{color}
\usepackage{adjustbox}
\usepackage{algorithm}
\usepackage{algorithmic}
\usepackage{multirow}
\usepackage{lipsum}
\usepackage{paralist}
\usepackage[table]{xcolor}
\usepackage{enumitem}

\usepackage{caption}
\usepackage{subcaption}
\usepackage{comment}

\usepackage[accsupp]{axessibility}  

\newcommand\norm[1]{\lVert#1\rVert}

\usepackage[breaklinks=true,bookmarks=false]{hyperref}

\iccvfinalcopy 

\def\iccvPaperID{****} 

\begin{document}

\title{FlipNeRF: Flipped Reflection Rays for Few-shot Novel View Synthesis}

\author{Seunghyeon Seo~~~
Yeonjin Chang~~~
Nojun Kwak
\smallskip
\\
Seoul National Univeristy\\
{\tt\small \{zzzlssh, yjean8315, nojunk\}@snu.ac.kr}
}

\maketitle

\begin{abstract}
   Neural Radiance Field (NeRF) has been a mainstream in novel view synthesis with its remarkable quality of rendered images and simple architecture.
   Although NeRF has been developed in various directions improving continuously its performance, the necessity of a dense set of multi-view images still exists as a stumbling block to progress for practical application.
   In this work, we propose FlipNeRF, a novel regularization method for few-shot novel view synthesis by utilizing our proposed flipped reflection rays.
   The flipped reflection rays are explicitly derived from the input ray directions and estimated normal vectors, and play a role of effective additional training rays while enabling to estimate more accurate surface normals and learn the 3D geometry effectively.
   Since the surface normal and the scene depth are both derived from the estimated densities along a ray, the accurate surface normal leads to more exact depth estimation, which is a key factor for few-shot novel view synthesis.
   Furthermore, with our proposed Uncertainty-aware Emptiness Loss and Bottleneck Feature Consistency Loss, FlipNeRF is able to estimate more reliable outputs with reducing floating artifacts effectively across the different scene structures, and enhance the feature-level consistency between the pair of the rays cast toward the photo-consistent pixels without any additional feature extractor, respectively.
   Our FlipNeRF achieves the SOTA performance on the multiple benchmarks across all the scenarios.
   The codes and more qualitative results are available in our project page: \url{https://shawn615.github.io/flipnerf/}.
\end{abstract}

\section{Introduction}
\label{sec:intro}

\begin{figure}[t]
    \centering
    \vspace{-.15cm}
    \begin{subfigure}[b]{\linewidth}
         \centering
        \includegraphics[width=\linewidth]{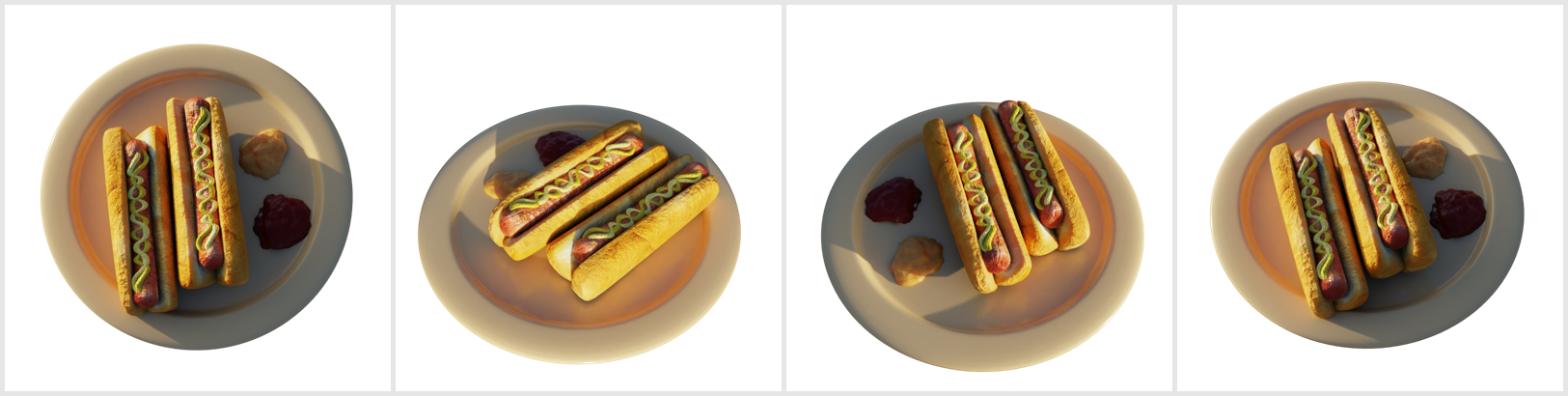}
        \caption{Sparse training views}
     \end{subfigure}
    \begin{tabular}{p{0.1\textwidth}p{0.093\textwidth}p{0.09\textwidth}p{0.1\textwidth}}
    \midrule
     \centering\scriptsize mip-NeRF~\cite{barron2021mip} & \centering\scriptsize Ref-NeRF~\cite{verbin2022ref} & \centering\scriptsize MixNeRF~\cite{seo2023mixnerf} & \centering\scriptsize FlipNeRF (Ours)
    \end{tabular}
    \begin{subfigure}[b]{\linewidth}
         \centering
        \includegraphics[width=\linewidth]{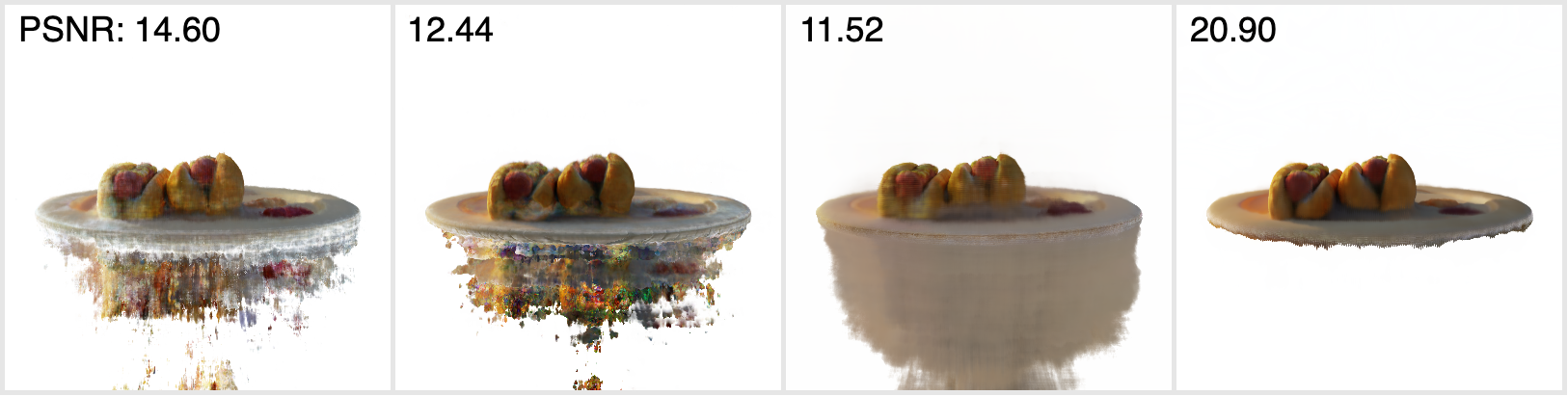}
    \end{subfigure}
    \begin{subfigure}[b]{\linewidth}
         \centering
        \includegraphics[width=\linewidth]{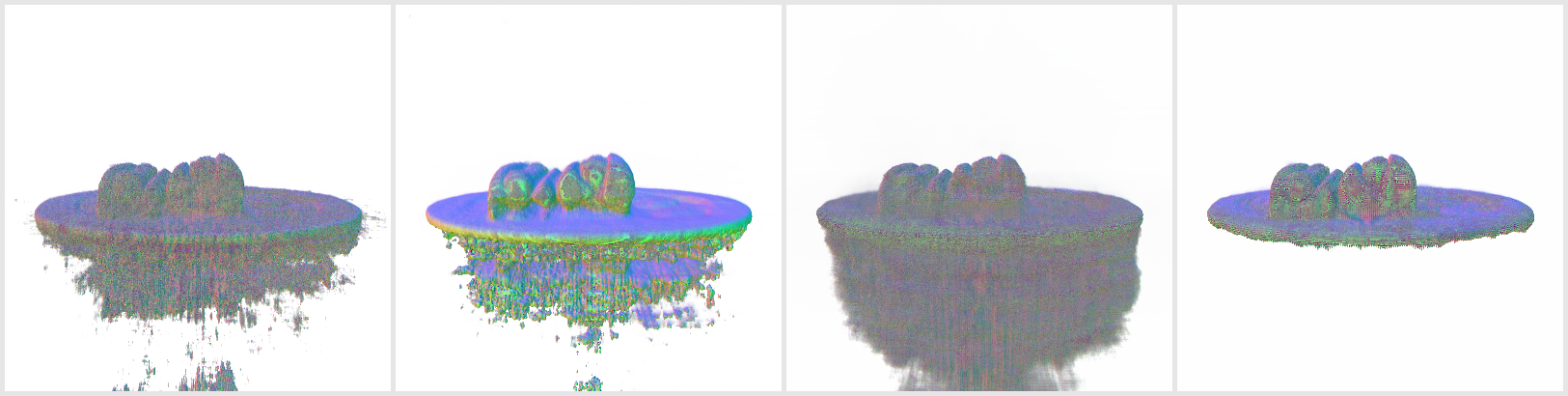}
    \end{subfigure}
    \begin{subfigure}[b]{\linewidth}
         \centering
        \includegraphics[width=\linewidth]{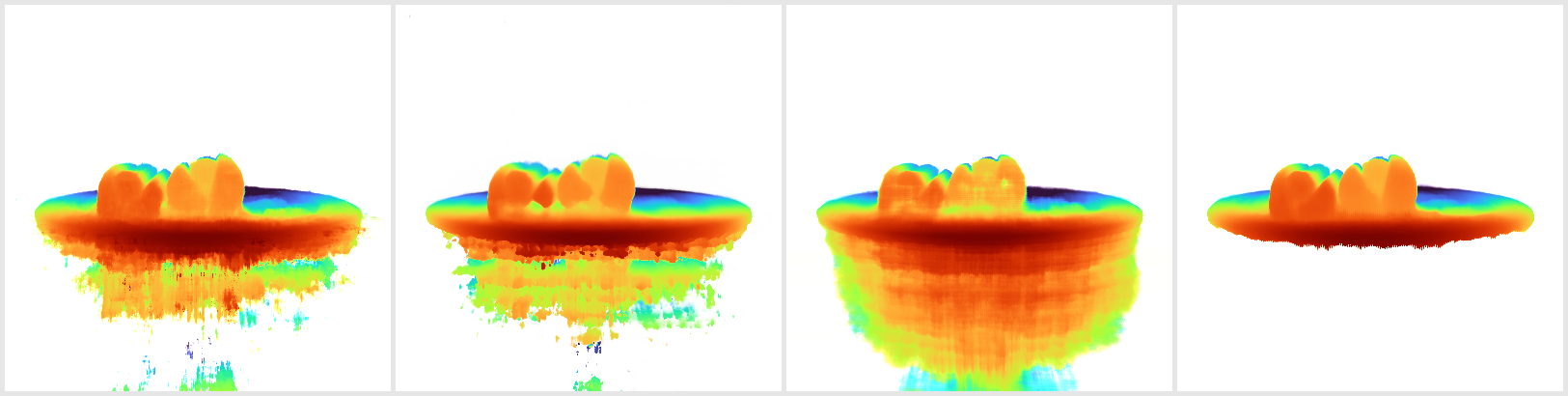}
        \caption{Synthesized unseen views}
     \end{subfigure}
     \vspace{-6mm}
    \caption{
    \textbf{Synthesis results from sparse inputs.}
    Our FlipNeRF significantly improves rendering quality compared to other baselines.
    Compared to the vanilla mip-NeRF~\cite{barron2021mip} and MixNeRF~\cite{seo2023mixnerf}, which is the state-of-the-art regularization method, ours reduces the noises and floating artifacts noticeably with superior surface normal estimation.
    Although Ref-NeRF~\cite{verbin2022ref} estimates smooth normal vectors, it shows much inferior rendering results to ours with a large chunk of noise under the few-shot setting.
    }
    \label{fig:intro}
    \vspace{-3mm}
\end{figure}

Neural Radiance Field (NeRF)~\cite{mildenhall2021nerf} has achieved great success in rendering photo-realistic images from novel viewpoints.
However, the necessity of a dense set of training images remains as a practical bottleneck since it suffers from significant performance degradation when trained with sparse views.

There are two mainstreams for few-shot novel view synthesis: \textit{pre-training} and \textit{regularization} methods, both of which focus on learning the 3D geometry efficiently from sparse inputs.
The pre-training methods~\cite{yu2021pixelnerf, chen2021mvsnerf, chibane2021stereo, wang2021ibrnet, liu2022neural, jang2021codenerf, li2021mine, rematas2021sharf, trevithick2021grf, johari2022geonerf} require large-scale datasets consisting of different scenes with multi-view images for injecting prior knowledge during the pre-training, while the regularization methods~\cite{niemeyer2022regnerf, seo2023mixnerf, kim2022infonerf, jain2021putting, roessle2022dense, deng2022depth, kwak2023geconerf} are optimized per scene, exploiting additional training resources, \eg unseen viewpoints~\cite{niemeyer2022regnerf, kim2022infonerf, kwak2023geconerf}, depth map generation~\cite{deng2022depth, roessle2022dense}, off-the-shelf models~\cite{jain2021putting, niemeyer2022regnerf}, and so on, for an effective regularization to alleviate overfitting.
Although the prior arts achieved promising results in novel view synthesis from sparse inputs, there still exist hurdles to overcome.
The large-scale datasets, which are used for pre-training methods, are expensive to collect and the NeRF model is prone to performance degradation for the out-of-distribution dataset.
On the other hand, the regularization methods heavily rely on additional training resources which might not always be available and require many heuristic factors, \eg the choice of off-the-shelf models, the hyperparameters for sampling unseen viewpoints, and so on.

In this paper, we propose \textit{FlipNeRF}, which is an effective regularization method exploiting the flipped reflection rays\footnote{The term `flipped' is used because the reflected ray has an opposite direction (from an object to a camera).} as additional training resources with filtering the ineffective newly generated rays.
We derive a batch of flipped reflection rays from the original ray directions and estimated surface normals so that they are cast toward the same target pixels of the original input ray.
Compared to the existing regularization methods which have mainly focused on the accurate depth estimation from limited input views~\cite{seo2023mixnerf, niemeyer2022regnerf, kim2022infonerf, roessle2022dense, deng2022depth}, our FlipNeRF is trained to reconstruct surface normals accurately by learning to generate effective reflection rays to be used in training.
Since both estimated surface normals and depths are derived from the volume densities representing underlying 3D geometry, accurately estimating the surface normals of an object naturally leads to more accurate depth maps.

Furthermore, we propose an effective regularization loss, \textit{Uncertainty-aware Emptiness Loss (UE Loss)}, to reduce the floating artifacts effectively while considering the uncertainty of the model's outputs by using the estimated scale parameters for mixture models.
Since our FlipNeRF is built upon MixNeRF~\cite{seo2023mixnerf}, which is a regularization method achieving promising results by modeling input rays with mixture density models~\cite{bishop1994mixture}, we are able to apply our proposed loss without any modification of the architecture by using the estimated scale parameters of each sample along a ray, which stand for the uncertainty of the samples' estimated probability density distributions.

Additionally, inspired by \cite{chibane2021stereo, jain2021putting, kim2022infonerf} which address the feature-level consistency of targets under the sparse input setting, we encourage the consistency for the pairs of bottleneck features between the original input rays and flipped reflection rays.
We leverage a Jensen-Shannon Divergence, which is based on the similarity between the probability distributions, to make the pairs of bottleneck feature distributions of original and flipped reflection rays more similar to each other improving feature consistency.

We demonstrate the effectiveness of our proposed FlipNeRF through the experiments on the multiple benchmarks, \eg Realistic Synthetic 360$^\circ$~\cite{mildenhall2021nerf}, DTU~\cite{jensen2014large}, and LLFF~\cite{mildenhall2019local}.
Our method achieves state-of-the-art (SOTA) performances compared to other baselines.
Especially, ours outperforms other baselines by a large margin with more accurate surface normals under the extremely sparse settings such as 3/4-view setting which are the most challenging ones.
Our contributions are summarized as follows:

\begin{itemize}[noitemsep,topsep=0pt,parsep=0pt,partopsep=0pt, leftmargin=*] 
    \item We propose an effective training framework for NeRF with sparse training views, called \textit{FlipNeRF}. It leverages flipped reflection rays to provide additional training resources, resulting in more precise surface normals with our proposed masking strategy to filter the ineffective rays.
    \item We also propose an effective regularization loss, \textit{Uncertainty-aware Emptiness Loss (UE Loss)}, which reduces floating artifacts with considering the uncertainty of outputs, leading to more reliable estimation.
    \item We enhance the consistency of bottleneck features between the original input rays and flipped reflection rays by Jensen-Shannon Divergence, coined as \textit{Bottleneck Feature Consistency Loss (BFC Loss)}, improving the robustness for rendering from unseen viewpoints.
    \item Our FlipNeRF achieves SOTA performance over the multiple benchmarks.
    Especially, ours outperforms other baselines by a large margin in more challenging scenarios, \eg 3/4-view.
\end{itemize}

\begin{figure*}[t]
\centering
\includegraphics[width=0.9\linewidth]{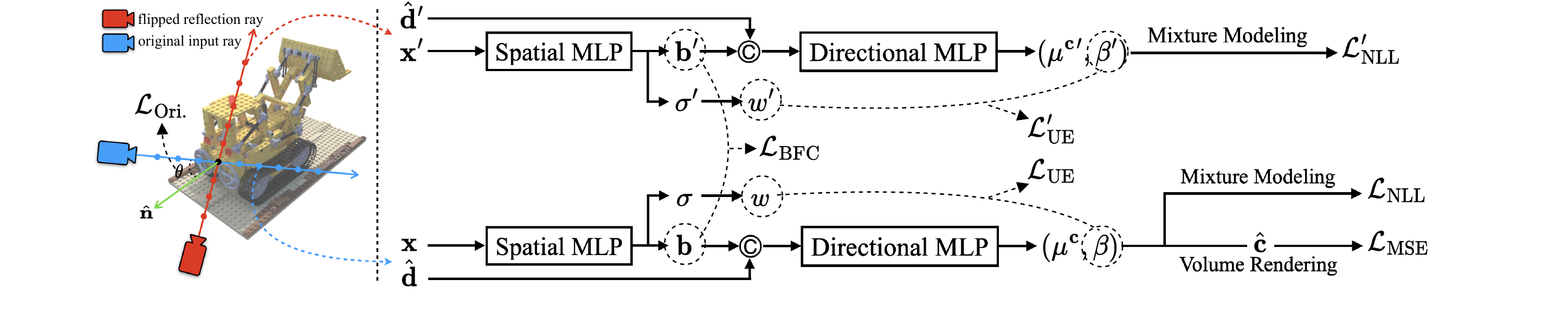}
\caption{\textbf{Overall framework of FlipNeRF.}
Our FlipNeRF utilize the newly generated flipped reflection rays with our proposed UE Loss and BFC Loss as well as existing MSE, NLL and Orientation losses.
See Sec.~\ref{sec:method} and Fig.~\ref{fig:ref_gen} for more details about generation process of flipped reflection rays and the loss terms.
}
\vspace{-.3cm}
\label{fig:overview}
\end{figure*}

\section{Related Works}
\label{sec:rel_works}
\subsection{Neural Radiance Field}
\label{subsec:nerf}
Recently, Neural Radiance Field (NeRF) \cite{mildenhall2021nerf} has shown impressive performance and potential in the novel view synthesis task.
NeRF represents a scene with an MLP, mapping coordinates and viewing directions to its colors and volume density, and then creates a novel view through volume rendering.
Subsequent studies have developed NeRF in several directions, \eg using conical frustums instead of rays~\cite{barron2021mip}, reparameterizing an input viewing direction as its reflection direction~\cite{verbin2022ref}, and so on. 
These works have made significant progress by addressing the various issues in novel view synthesis, but there still exists a limitation in that NeRF requires a dense set of training images and a lengthy training time.
Many studies have addressed these issues \cite{garbin2021fastnerf, reiser2021kilonerf, chen2021mvsnerf}, including the utilization of various data structures for faster training and inference \cite{fridovich2022plenoxels, yu2021plenoctrees} and attempts to train NeRF with only a few training images.
Our work focuses on enhancing the performance of NeRF when a sparse set of views are provided as training images.

\subsection{Few-Shot Novel View Synthesis}
\label{subsec:fewshot_nerf}
There are two main approaches for a few-shot novel view synthesis: the \textit{pre-training} and the \textit{regularization} method.
The pre-training methods require a large dataset of multi-view scenes to provide prior knowledges of 3D geometry to a NeRF model and then optionally finetune on the target scene~\cite{chen2021mvsnerf, chibane2021stereo, jang2021codenerf, li2021mine, rematas2021sharf, trevithick2021grf, wang2021ibrnet, yu2021pixelnerf}.
On the contrary, the regularization methods~\cite{seo2023mixnerf, kim2022infonerf, jain2021putting, niemeyer2022regnerf, deng2022depth, roessle2022dense, kwak2023geconerf} are optimized per scene without pre-training process by exploiting additional training resources, \eg depth maps \cite{deng2022depth, roessle2022dense} and semantic consistency \cite{jain2021putting}, as an extra supervision.
Among them, \cite{niemeyer2022regnerf, kim2022infonerf, kwak2023geconerf} adopt an unseen viewpoint sampling strategy to make up for insufficient training views.
However, these sampling processes require many hand-designed factors such as the ranges of rotation, translation, jittering, and so on, which can introduce artificial biases.
Our work proposes a novel regularization approach to derive a set of flipped reflection rays from estimated surface normals and utilizes these for regularization, which does not require a heuristic factor to be finetuned, resulting in more effective training strategy with limited inputs.

\subsection{Surface Normal Reconstruction}
\label{subsec:surface_normal}
There is a line of research to recover accurate textures and lighting conditions of objects with NeRF~\cite{oechsle2021unisurf, yariv2021volume, wang2021neus, darmon2022improving, srinivasan2021nerv, zhang2021physg, boss2021nerd}.
Although these earlier studies successfully reconstruct the high-quality isosurfaces derived from the scene representations, their rendering quality for novel views is still inferior to the NeRF-like models.
Meanwhile, Ref-NeRF~\cite{verbin2022ref} achieved superior performance with remarkable quality of surface normals compared to the existing NeRF models.
Since the normal vectors utilized in NeRF framework are derived from the negative normalized density gradients~\cite{srinivasan2021nerv, boss2021nerd, verbin2022ref}, which represent the underlying geometry of 3D scenes, learning an accurate density distribution along a ray is a key factor for surface normal reconstruction.
However, for the few-shot novel view synthesis, the prior works mostly focus on the accurate depth estimation without attention to the surface normals, both of which are derived from the estimated volume densities.
In this work, we approach the few-shot novel view synthesis problem with focusing on the surface normals, which is another critical factor for an effective learning of 3D scene geometry.
To the best of our knowledge, our work is the first attempt to focus on the surface normal estimation for few-shot novel view synthesis.

\section{Method}
\label{sec:method}
In this work, we propose an effective regularization method for few-shot novel view synthesis with flipped reflection rays.
Our FlipNeRF is built upon MixNeRF~\cite{seo2023mixnerf} which leverages a mixture model framework (Sec.~\ref{subsec:preliminary}).
We derive a batch of flipped reflection rays and cast them toward the identical target pixels as additional training rays (Sec.~\ref{subsec:ref_ray}).
Furthermore, we propose the \textit{Uncertainty-aware Emptiness Loss} and \textit{Bottleneck Feature Consistency Loss} to alleviate the floating artifacts adaptively based on the uncertainty and enhance the consistency between the bottleneck feature distributions of the original and flipped reflection rays, respectively (Sec.~\ref{subsec:ue_loss} and Sec.~\ref{subsec:feat_con}).
Finally, our FlipNeRF is trained to minimize the MSE and NLL losses as well as the proposed regularization loss terms with their corresponding balancing weights (Sec.~\ref{subsec:total_loss}).
Fig.~\ref{fig:overview} shows an overview of our FlipNeRF.

\begin{figure}[!t]
\centering
\includegraphics[width=0.8\linewidth]{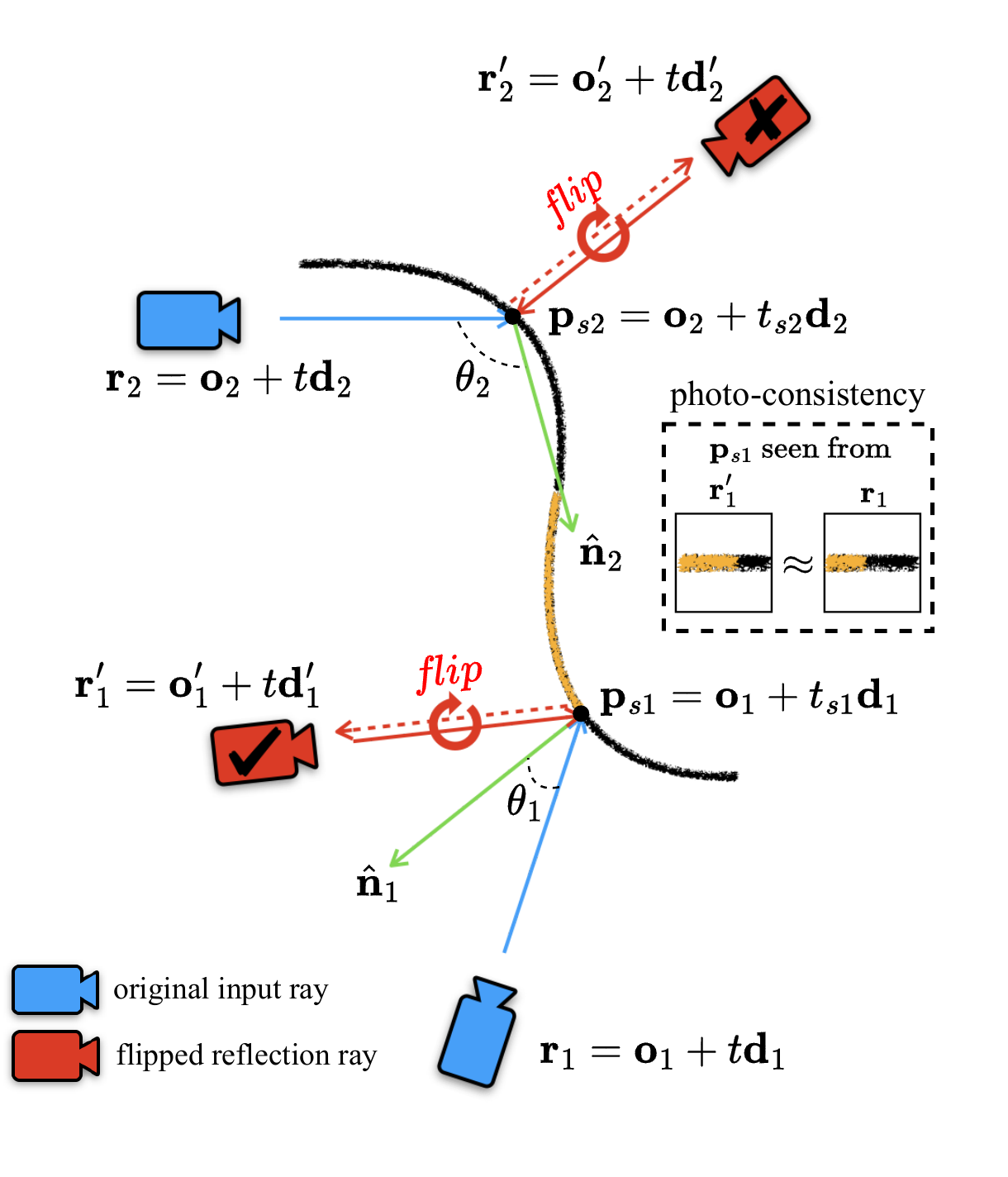}
\vspace{-5mm}
\caption{\textbf{Flipped reflection ray generation.}
        Our FlipNeRF generates the flipped reflection ray $\mathbf{r}'$ from the estimated normal vector $\hat{\mathbf{n}}$ and original input ray direction $\mathbf{d}$.
        With our masking strategy, $\mathbf{r}_{2}'$ is filtered out since it does not satisfy the photo-consistency condition, \ie $\theta_2$ is bigger than 90$^\circ$.
        The smaller $\theta$ is, the more photo-consistent the target pixel is, where the pair of $\mathbf{r}$ and $\mathbf{r}'$ are cast.}
\vspace{-.3cm}
\label{fig:ref_gen}
\end{figure}

\subsection{Preliminaries}
\label{subsec:preliminary}
\paragraph{NeRF.}
The NeRF~\cite{mildenhall2021nerf}, which is an MLP-based neural network, represents a 3D scene as a continuous radiance field of RGB color and volume density.
For every point sampled along a ray, the 3D coordinates $\mathbf{x} = (x, y, z)$ and viewing directions $(\theta, \phi)$ are mapped to the colors $\mathbf{c} = (r, g, b)$ and densities $\sigma$:
\begin{equation}
  F(\gamma(\mathbf{x}), \gamma(\hat{\mathbf{d}})) \to (\mathbf{c}, \sigma),
  \label{eq:nerf}
\end{equation}
where $F(\cdot)$, $\gamma(\cdot)$, and $\hat{\mathbf{d}}$ indicate an MLP, the positional encoding for the inputs, and the 3D Cartesian unit vector used as an input viewing direction in practice, respectively.

The volumetric radiance field is rendered by alpha compositing the RGB values along an input ray $\mathbf{r}(t) = \mathbf{o} + t\mathbf{d}$~\cite{max1995optical}, where $\mathbf{o}$ and $\mathbf{d}$ denote the camera origin and unnormalized direction vector, \ie $\mathbf{d} = \norm{\mathbf{d}}_2 \cdot \hat{\mathbf{d}}$, respectively.
The volume rendering integrals are denoted as follows:
\begin{equation}
  \mathbf{\hat{c}(r)} = \int_{t_n}^{t_f} T(t)\sigma(\mathbf{r}(t))\mathbf{c}(\mathbf{r}(t), \hat{\mathbf{d}}) \,dt,
\end{equation}
\label{eq:integral}
where $T(t) = \exp (-\int_{t_n}^{t} \sigma(s) \,ds)$ indicates the degree of transparency.
In practice, it is approximated with numerical quadrature~\cite{mildenhall2021nerf} by sampling points along a ray.

The radiance field is trained to minimize the mean squared error (MSE) between rendered and GT pixels:
\begin{equation}
  \mathcal{L}_\text{MSE} = \sum_{\mathbf{r} \in \mathcal{R}} || \mathbf{\hat{c}(r)} - \mathbf{c^\text{GT}(r)} ||^2_2\,,
  \label{eq:mse}
\end{equation}
where $\mathcal{R}$ is denoted as a set of rays.

\paragraph{MixNeRF.}
Built upon mip-NeRF~\cite{barron2021mip}, which leveraged a cone tracing method and proposed an integrated positional encoding to address an aliasing problem, MixNeRF~\cite{seo2023mixnerf} estimates the joint probability distribution of color values and models a ray with a mixture of densities:
\begin{equation}
p(\mathbf{c}|\mathbf{r}) = \sum_{i=1}^{M} \pi_{i}\mathcal{F}(\mathbf{c}; \mu_{i}^\mathbf{c}, \beta_{i}),
\label{eq:mixture_pdf}
\end{equation}
where $M$ is the number of sampled points, $\mathcal{F}(\mathbf{c}; \mu_{i}^\mathbf{c}, \beta_{i})$ denotes the Laplacian distribution of RGB $\mathbf{c}$ with location parameter $\mu_{i}^\mathbf{c} \in \{\mu_{i}^r, \mu_{i}^g, \mu_{i}^b\}$, \ie estimated RGB values of sample, and scale parameter $\beta_{i} \in \{\beta_{i}^r, \beta_{i}^g, \beta_{i}^b\}$.
The mixture coefficient $\pi_{i}$ is derived from the estimated volume density $\sigma_{i}$ as follows:
\begin{equation}
\pi_{i} = \frac{w_i}{\sum_{m=1}^{M} w_m} = \frac{T_i(1-\exp(-\sigma_{i}\delta_{i}))}{\sum_{m=1}^{M} T_m(1-\exp(-\sigma_{m}\delta_{m}))},
\label{eq:mix_coef}
\end{equation}
where $w_i$ and $\delta_{i}$ indicate the alpha blending weight and sample interval, respectively.
Thanks to the mixture model's capacity of representing complex distributions, MixNeRF learns the density distribution effectively with sparse inputs by minimizing the negative log-likelihood (NLL) in Eq.~\ref{eq:mixture_pdf}.

Our FlipNeRF is built upon MixNeRF leveraging the mixture modeling framework while achieving superior rendering quality with noticeably fewer artifacts and more accurate surface normals to MixNeRF.

\subsection{Auxiliary Flipped Reflection Ray}
\label{subsec:ref_ray}
As shown in Fig.~\ref{fig:ref_gen}, we exploit a batch of flipped reflection rays $\mathbf{r}' \in \mathcal{R}'$ as extra training resources, which are derived from the original input ray directions $\mathbf{d}$ and estimated surface normals $\hat{\mathbf{n}}$.
First, we derive a flipped reflection direction $\mathbf{d}'$ from $\mathbf{d}$ and $\hat{\mathbf{n}}$:
\begin{equation}
\mathbf{d}' = 2(\mathbf{d} \cdot \hat{\mathbf{n}})\hat{\mathbf{n}} - \mathbf{d},
\label{eq:ref}
\end{equation}
where $\hat{\mathbf{n}}$ denotes the weighted sum of blending weights and estimated normal vectors along a ray, \ie $\hat{\mathbf{n}} = \sum_{i=1}^{M} w_{i}\mathbf{n}_{i}$.
\footnote{Technically, $\hat{\mathbf{n}}$ is not guaranteed to be a unit vector without an explicit normalization process.
However, we empirically found that the normalization rather destabilizes the training and leads to the performance degradation.
Kindly refer to our supplementary material for related experiments.}
Note that we use the gradient of volume density as estimated surface normals following~\cite{boss2021nerd, srinivasan2021nerv, verbin2022ref}.

To generate the additional training rays based on $\mathbf{d}'$, we need a set of imaginary ray origins $\mathbf{o}'$ located in a suitable space considering the hitting point and the original input ray origins $\mathbf{o}$.
Since the vanilla NeRF models, which are trained with a dense set of images, tend to have the blending weight distribution whose peak is located on the point around the object surface $\mathbf{p}_s = \mathbf{o} + t_s\mathbf{d}$~\cite{deng2022depth, seo2023mixnerf}, \ie the $s$-th sample whose blending weight is the highest along a ray.
Therefore, we place $\mathbf{o}'$ so that the $s$-th sample of $\mathbf{r}'$ is $\mathbf{p}_s$:
\begin{equation}
\mathbf{o}' = \mathbf{p}_s - t_s\mathbf{d}',
\label{eq:ref_cam_org}
\end{equation}
resulting in our proposed flipped reflection ray, $\mathbf{r}'(t) = \mathbf{o}' + t\mathbf{d}'$.
Compared to the previous unseen viewpoint sampling strategies~\cite{kim2022infonerf, niemeyer2022regnerf, kwak2023geconerf}, our proposed strategy does not rely on the randomness of unseen viewpoint sampling and reduces the heuristic factors for sampling schemes, \eg the range of rotation, translation, and so on.
Furthermore, since our newly generated $\mathbf{r}'$ are cast on the identical object surfaces where the original input rays $\mathbf{r}$ are cast, \ie the target pixels are photo-consistent for the pair of $\mathbf{r}$ and $\mathbf{r}'$ without any sophisticated viewpoint sampling process, we are able to train $\mathbf{r}'$ effectively with the same GT pixels of $\mathbf{r}$.

However, since $\hat{\mathbf{n}}$, which are used to derive $\mathbf{d}'$, are not the ground truth but the estimation, there exists a concern that even miscreated $\mathbf{r}'$, which do not satisfy photo-consistency, can be used for training.
As a result, it might lead to performance degradation while providing misleading training cues.
To address this problem, we mask the ineffective $\mathbf{r}'$ by considering the angle $\theta$ between $\hat{\mathbf{n}}$ and $-\hat{\mathbf{d}}$ as follows:
\begin{equation}
M(\mathbf{r}') = 
\begin{cases}
1  & \text{if~} \arccos{(-(\hat{\mathbf{d}} \cdot \hat{\mathbf{n}}))} < \tau \\
0  & \text{otherwise}
\end{cases},
\label{eq:masking}
\end{equation}
where $-(\hat{\mathbf{d}} \cdot \hat{\mathbf{n}})$ amounts to $\cos{\theta}$ of original input rays and normal vectors, and $\tau$ indicates the threshold for filtering the invalid rays, which we set as $90^\circ$ unless specified.
Through this masking process, only $\mathbf{r}'$ which are cast toward the photo-consistent point can be remained as we intend.
Finally, our proposed flipped reflection rays are modeled by mixture density like the original input rays:
\begin{equation}
p(\mathbf{c}|\mathbf{r}') = \sum_{i=1}^{M} \pi_{i}'\mathcal{F}(\mathbf{c}; \mu_{i}^\mathbf{c}{'}, \beta_{i}').
\label{eq:mixture_pdf2}
\end{equation}

Additionally, we leverage the \textit{Orientation Loss} proposed in Ref-NeRF~\cite{verbin2022ref} to penalize the backward-facing normal vectors for learning accurate surface normals:
\begin{equation}
l_{\text{Ori.}}(\mathbf{r}) = \sum_{i=1}^{M} w_i\max(0, \mathbf{n}_{i} \cdot \hat{\mathbf{d}})^2.
\label{eq:ori_loss}
\end{equation}
Unlike Ref-NeRF, we penalize underlying density gradient normal $\mathbf{n}_{i}$ instead of predicted normals.

Note that our FlipNeRF is fundamentally different compared to Ref-NeRF since ours generates additional training rays through the derivation of reflection direction without modification to original representations while Ref-NeRF replaced the input viewing direction with its reflection direction, reparameterizing the outgoing radiance.

\subsection{Uncertainty-aware Regularization}
\label{subsec:ue_loss}
Several regularization techniques have been proposed to reduce the floating artifacts present in synthesized images, which is one of the major problems of NeRF.
Among them, we leverage the \textit{Emptiness Loss}~\cite{wang2022score} which penalizes the small blending weights along a ray as follows:
\begin{equation}
l_{\text{Emp.}}(\mathbf{r}) = \frac{1}{M} \sum_{i=1}^{M} \log(1 + \eta \cdot w_i),
\label{eq:emp_loss}
\end{equation}
where the bigger $\eta$ is, the steeper the loss function becomes around $0$.

However, the naive application of existing regularization techniques with limited training views might not be consistently helpful across the different scenes due to the scene-by-scene different structure, resulting in overall performance degradation.
To address this problem, we propose \textit{Uncertainty-aware Emptiness Loss (UE Loss)} developed upon the Emptiness Loss, which reduces the floating artifacts consistently over the different scenes by considering the output uncertainty:
\begin{equation}
\begin{split}
l_{\text{UE}}(\mathbf{r}) = \frac{1}{M} \sum_{i=1}^{M} \log(1 + \rho \cdot \eta \cdot w_i), \\
    \text{where} \quad \rho = \frac{1}{3} \sum_{c}^{\{r,g,b\}} \sum_{i=1}^{M} \beta_i^c .
\label{eq:ue_loss}
\end{split}
\end{equation}
$\rho$ amounts to the average of the summation of estimated scale parameters of RGB color distributions from all samples along a ray, which we use as the uncertainty of a ray.
By our proposed UE Loss, we are able to regularize the blending weights adaptively, \ie the more uncertain a ray is, the more penalized the blending weights along the ray are.
It is able to reduce floating artifacts consistently across the scenes with different structures and enables to synthesize more reliable outputs by considering uncertainty.

\subsection{Bottleneck Feature Consistency}
\label{subsec:feat_con}
Motivated by previous works addressing the feature-level consistency of multiple views for few-shot novel view synthesis~\cite{chibane2021stereo, jain2021putting, kim2022infonerf}, we encourage the consistency of bottleneck feature distributions between $\mathbf{r}$ and $\mathbf{r}'$, which are intermediate feature vectors, \ie outputs of the spatial MLP of NeRF, by Jensen-Shannon Divergence (JSD):
\begin{equation}
l_{\text{BFC}}(\mathbf{r}, \mathbf{r}') = JSD(\psi(\mathbf{b}), \psi(\mathbf{b}')),
\label{eq:con_loss}
\end{equation}
where $\psi(\cdot)$, $\mathbf{b}$ and $\mathbf{b}'$ denote the softmax function, the bottleneck features of $\mathbf{r}$ and $\mathbf{r}'$, respectively.
While the existing methods~\cite{chibane2021stereo, jain2021putting} rely on off-the-shelf feature extractors like 2D CNN or CLIP~\cite{radford2021learning} to address high-level feature consistency, we regulate the pair of features effectively by enhancing consistency between bottleneck features without depending on additional feature extractors.

\subsection{Total Loss}
\label{subsec:total_loss}
Our FlipNeRF is not only trained to maximize the log-likelihood of the target pixel $\mathbf{c}_\text{GT}$ for a set of original input rays $\mathcal{R}$, but also for flipped reflection rays $\mathcal{R}_{M}'$, where the ineffective rays are excluded from the total flipped reflection rays $\mathcal{R}'$ by our masking strategy in Eq.~\ref{eq:masking}.
Likewise, the UE Losses are applied for both $\mathcal{R}$ and $\mathcal{R}_{M}'$.

Aggregating all, our total loss over a batch is as follows:
\begin{equation}
\begin{split}
\mathcal{L}_\text{Total} &= \mathcal{L}_\text{MSE} + \lambda_{1}\mathcal{L}_\text{NLL} + \lambda_{2}\mathcal{L}_\text{NLL}' \\
&+ \lambda_{3}\mathcal{L}_\text{UE} + \lambda_{4}\mathcal{L}_\text{UE}' + \lambda_{5}\mathcal{L}_\text{BFC} + \lambda_{6}\mathcal{L}_\text{Ori.},
\label{eq:total_loss}
\end{split}
\end{equation}
where a set of $\lambda$'s are balancing weight terms for the losses.

\section{Experiments}
\label{sec:exp}
\subsection{Experimental Settings}
\label{subsec:exp_setting}
\paragraph{Implementation details.}
Our FlipNeRF is built upon MixNeRF~\cite{seo2023mixnerf} based on the JAX codebase~\cite{bradbury2018jax}, which is developed upon mip-NeRF~\cite{barron2021mip}.
The overall training scheme follows~\cite{niemeyer2022regnerf, seo2023mixnerf}.
We adopt a scene space annealing strategy for the early training phase.
Also, we apply the exponential decay and warm up for the learning rate.
The Adam optimizer~\cite{kingma2014adam} and the gradient clippings by value at 0.1 and norm at 0.1 are used.
We train our FlipNeRF for 500 pixel epochs with a batch size of 4,096 on 4 NVIDIA RTX 3090.
Additionally, since the LLFF dataset~\cite{mildenhall2019local} consists of scenes with much more static movement of viewpoints compared to other datasets, we set the threshold ($\tau$) of our masking strategy as 30$^\circ$ for the experiments on LLFF.
More detailed hyperparameters and our loss balancing terms by the datasets and the number of training views are provided in the supplementary material.

\begin{table}[t]
\centering
\resizebox{\linewidth}{!}{
\begin{tabular}{l|c|c|c|c}
\toprule
  $\tau$ & PSNR $\uparrow$ & SSIM $\uparrow$ & LPIPS $\downarrow$ & Average Err. $\downarrow$\\
\midrule
30$^\circ$ & 18.62 & 0.747 & 0.206 & 0.121 \\
60$^\circ$ & 18.12 & 0.723 & 0.237 & 0.126 \\
90$^\circ$ & \cellcolor{red!50}\textbf{19.55} & \cellcolor{red!50}\textbf{0.767} & \cellcolor{red!50}\textbf{0.180} & \cellcolor{red!50}\textbf{0.101} \\
180$^\circ$ (No masking) & \cellcolor{red!25}\underline{18.76} & \cellcolor{red!25}\underline{0.755} & \cellcolor{red!25}\underline{0.190} & \cellcolor{red!25}\underline{0.111} \\
\bottomrule
\end{tabular}}
\vspace{-2mm}
\caption{
    \textbf{Comparison of masking conditions.}
    Our masking strategy with $\tau$ of 90$^\circ$ achieves the best results, filtering out the ineffective flipped reflection rays successfully.
    }
    \label{tab:anal_refrays}
\vspace{-3mm}
\end{table}

\begin{figure}[t]
     \centering
     \begin{subfigure}[b]{0.49\linewidth}
         \centering
         \includegraphics[width=\linewidth]{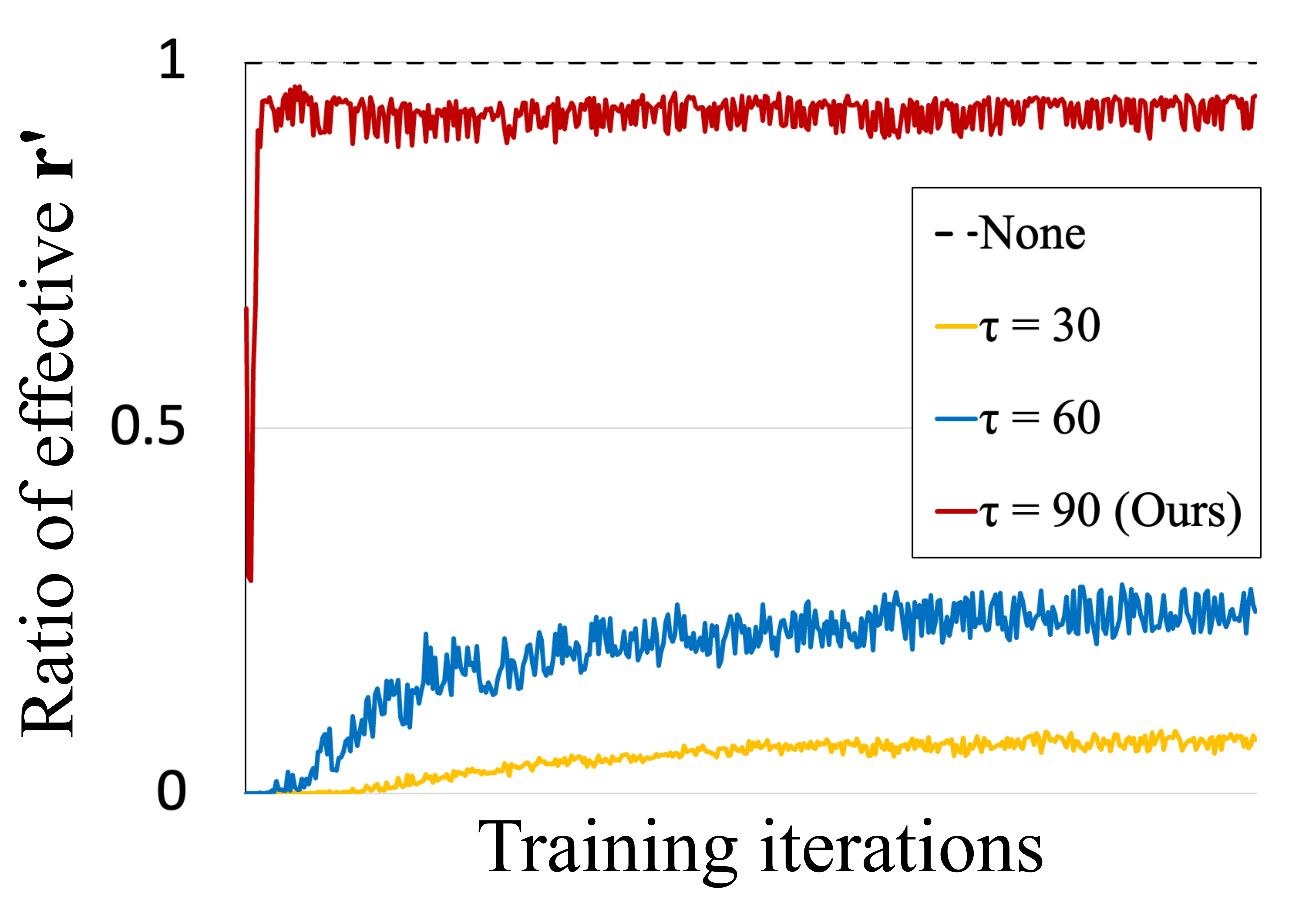}
         \vspace{-6mm}
         \caption{Effective $\mathbf{r}'$ ratio by $\tau$}
         \label{fig:mask_angle}
     \end{subfigure}
     \hfill
     \begin{subfigure}[b]{0.49\linewidth}
         \centering
         \includegraphics[width=0.9\linewidth]{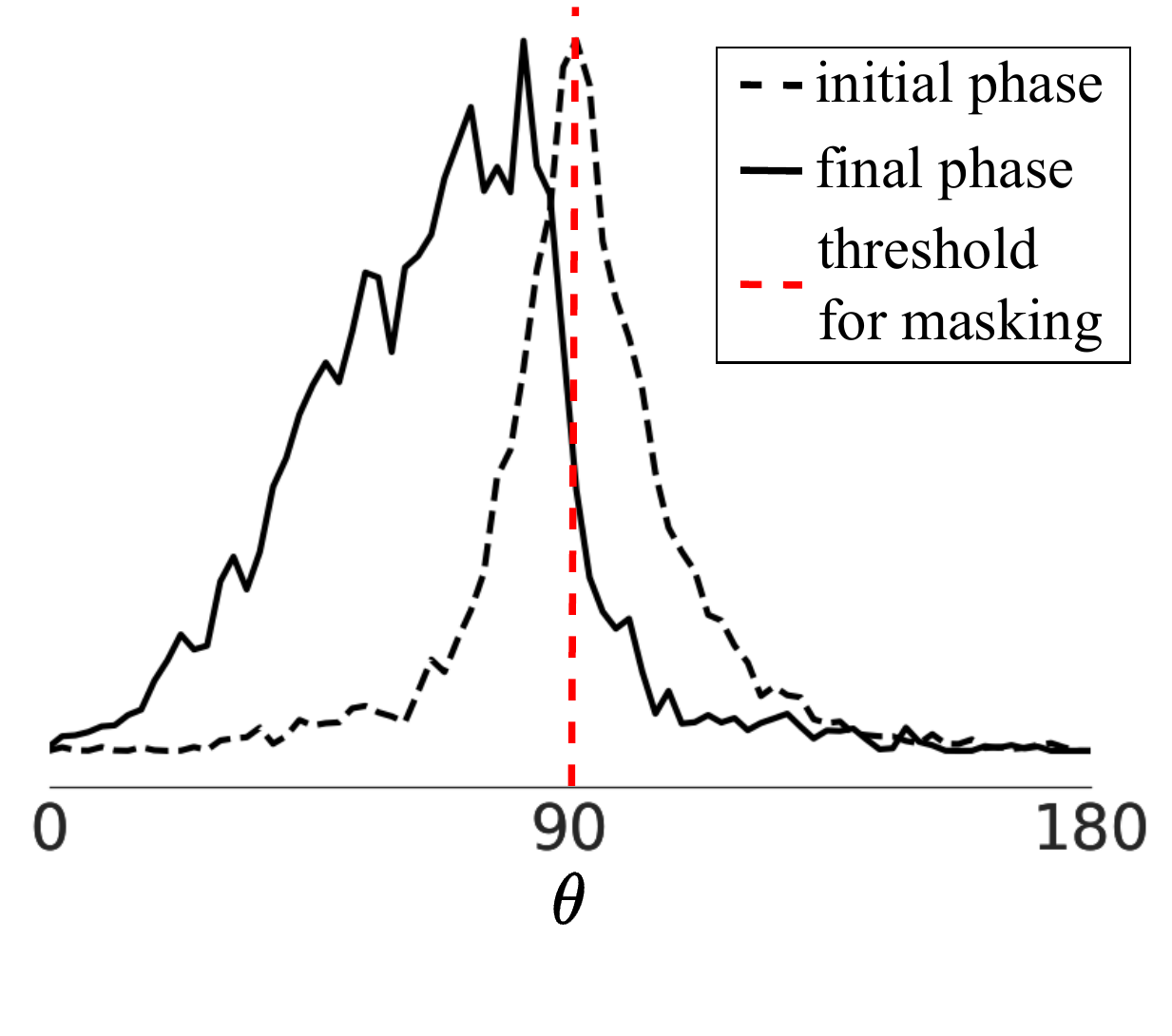}
         \vspace{-5mm}
         \caption{Distribution of $\theta$}
         \label{fig:flip_ref_ray_angle}
     \end{subfigure}
     \vspace{-2mm}
        \caption{\textbf{Analysis of the masking strategy.}
        }
    \vspace{-4mm}
        \label{fig:anal_refrays}
\end{figure}

\paragraph{Datasets and metrics.}
We evaluate the performance of our FlipNeRF and baselines on the representative benchmarks: Realistic Synthetic 360$^\circ$~\cite{mildenhall2021nerf}, DTU~\cite{jensen2014large}, and LLFF~\cite{mildenhall2019local}.
Realistic Synthetic 360$^\circ$ contains 8 synthetic scenes, each consisting of 400 multi-view rendered images with white background.
To compare against other representative baselines~\cite{kim2022infonerf, jain2021putting, seo2023mixnerf, niemeyer2022regnerf, barron2021mip}, we evaluate our FlipNeRF under the scenarios of 4 and 8 views.
For a fair comparison, we sample the first $n$-image of the training set for the $n$-view scenario so that the identical images are provided for training different methods following~\cite{seo2023mixnerf}. We use 200 images of the test set for evaluation.
For DTU, which provide various scenes including objects put on a white table with a black background, we conduct experiments on the 15 specific scenes under the scenarios of 3, 6, and 9-view, following the experimental protocol of \cite{yu2021pixelnerf}.
We also conduct a series of experiments for the analysis of FlipNeRF under 3-view setting as well as comparison against other baselines.
Additionally, we compare our FlipNeRF against other baselines on LLFF consisting of real forward-facing scenes, which is often tested as an out-of-distribution dataset for pre-training methods.
Following \cite{mildenhall2021nerf}, every 8th image of each scene is used for a held-out test set and the training views are evenly selected from the remaining images.
Like DTU, we report our results of the 3, 6, and 9-view scenarios, following~\cite{yu2021pixelnerf}.

For the quantitative evaluation for rendered images, we adopt the mean of PSNR, SSIM~\cite{wang2004image}, LPIPS~\cite{zhang2018unreasonable}, and the geometric average~\cite{barron2021mip}.
Furthermore, we also adopt the mean angular error (MAE$^\circ$)~\cite{verbin2022ref} and NLL~\cite{shen2021stochastic, shen2022conditional, loquercio2020general, lakshminarayanan2017simple} for evaluating the surface normals and uncertainty, respectively.
Specifically, following~\cite{shen2021stochastic}, we compute the NLL by deriving the probability of the GT pixel values given a Gaussian distribution with the estimated RGB values as mean and the uncertainty as variance, which is derived from the weighted sum of the blending weights and estimated scale parameters along a ray in our FlipNeRF.
For DTU, we report the results evaluated by masked metrics to prevent background bias, following \cite{niemeyer2022regnerf, seo2023mixnerf}.

\paragraph{Baselines.}
We compare our FlipNeRF against the SOTA regularization methods~\cite{jain2021putting, kim2022infonerf, niemeyer2022regnerf, seo2023mixnerf} on Realistic Synthetic 360$^\circ$ as well as the vanilla mip-NeRF~\cite{barron2021mip} and Ref-NeRF~\cite{verbin2022ref}, which is known for achieving promising results with accurate surface normals.
Furthermore, we compare ours against the representative pre-training methods~\cite{chen2021mvsnerf, chibane2021stereo, yu2021pixelnerf} as well as regularization methods on DTU and LLFF.
The pre-training baselines exploit the DTU and LLFF as pre-training dataset and out-of-distribution test set, respectively, while the regularization methods, mip-NeRF and Ref-NeRF are optimized per scene.
Note that we report the quantitative results of other baselines on DTU and LLFF from \cite{niemeyer2022regnerf}, which achieved better results than its original papers by the modified training scheme, and those on Realistic Synthetic 360$^\circ$ from \cite{seo2023mixnerf}, which trained the baselines with the identical training views for a fair comparison.

\subsection{Analysis of FlipNeRF}
\label{subsec:analysis}
\paragraph{Analysis of flipped reflection rays.}
As shown in Tab.~\ref{tab:anal_refrays}, our masking strategy of filtering out the ineffective flipped reflection rays with  the threshold ($\tau$) of 90$^\circ$ achieves the best performance among different options.
With $\tau$ of 30$^\circ$ and 60$^\circ$, the rendering quality is rather degraded since the newly generated rays are overly-filtered and do not provide enough additional supervision as demonstrated in Fig.~\ref{fig:mask_angle}.
On the other hand, when we exploit all the flipped reflection rays without masking, there exists a little improvement of performance compared to $30^\circ$ and $60^\circ$ masking, but it is still much inferior to $90^\circ$ due to the negative impact from the ineffective rays.
Additionally, Fig.~\ref{fig:flip_ref_ray_angle} shows the distribution of $\theta$, \ie angles between the input viewing directions and normal vectors.
As the smaller $\theta$ is, the more photo-consistent the target pixel is, \ie the more effective the newly generated flipped reflection rays are for training.
Since our FlipNeRF is trained to estimate the accurate normal vectors, we are able to exploit a set of more effective flipped reflection rays through the training, which are cast on the more photo-consistent target pixels.
Furthermore, at the initial training phase, the invalid additional rays are filtered effectively by our masking strategy, leading to high-quality supervision and stabilizing the training.

\begin{table}[t]
\centering
\resizebox{\linewidth}{!}{
\begin{tabular}{l|c|c|c|c|c}
\toprule
  & PSNR $\uparrow$ & SSIM $\uparrow$ & LPIPS $\downarrow$ & Average $\downarrow$ & NLL $\downarrow$\\
  \midrule
  MixNeRF~\cite{seo2023mixnerf} & 18.95 & 0.744 & 0.203 & 0.113 & 9.99 \\
\midrule
FlipNeRF (Ours) & & & & & \\
\quad w/o $\mathcal{L}_\text{Emp.}$ or $\mathcal{L}_\text{UE}$  & \cellcolor{red!25}\underline{19.30} & \cellcolor{red!25}\underline{0.758} & \cellcolor{red!25}\underline{0.196} & \cellcolor{red!25}\underline{0.108} & \cellcolor{red!25}\underline{4.88} \\
\quad w/ $\mathcal{L}_\text{Emptiness}$ & 18.62 & 0.749 & 0.204 & 0.118 & 4.92 \\
\quad w/ $\mathcal{L}_\text{UE}$ & \cellcolor{red!50}\textbf{19.55} & \cellcolor{red!50}\textbf{0.767} & \cellcolor{red!50}\textbf{0.180} & \cellcolor{red!50}\textbf{0.101} & \cellcolor{red!50}\textbf{2.56} \\
\bottomrule
\end{tabular}}
\vspace{-2mm}
\caption{
    \textbf{Effectiveness of $\mathcal{L}_\text{UE}$.}
    Our proposed $\mathcal{L}_\text{UE}$ improves the rendering quality consistently across the scenes with considering the uncertainty.
    }
    \label{tab:anal_ueloss}
\vspace{-2mm}
\end{table}

\begin{figure}[t]
    \centering
    \begin{subfigure}[b]{\linewidth}
    \includegraphics[width=\linewidth]{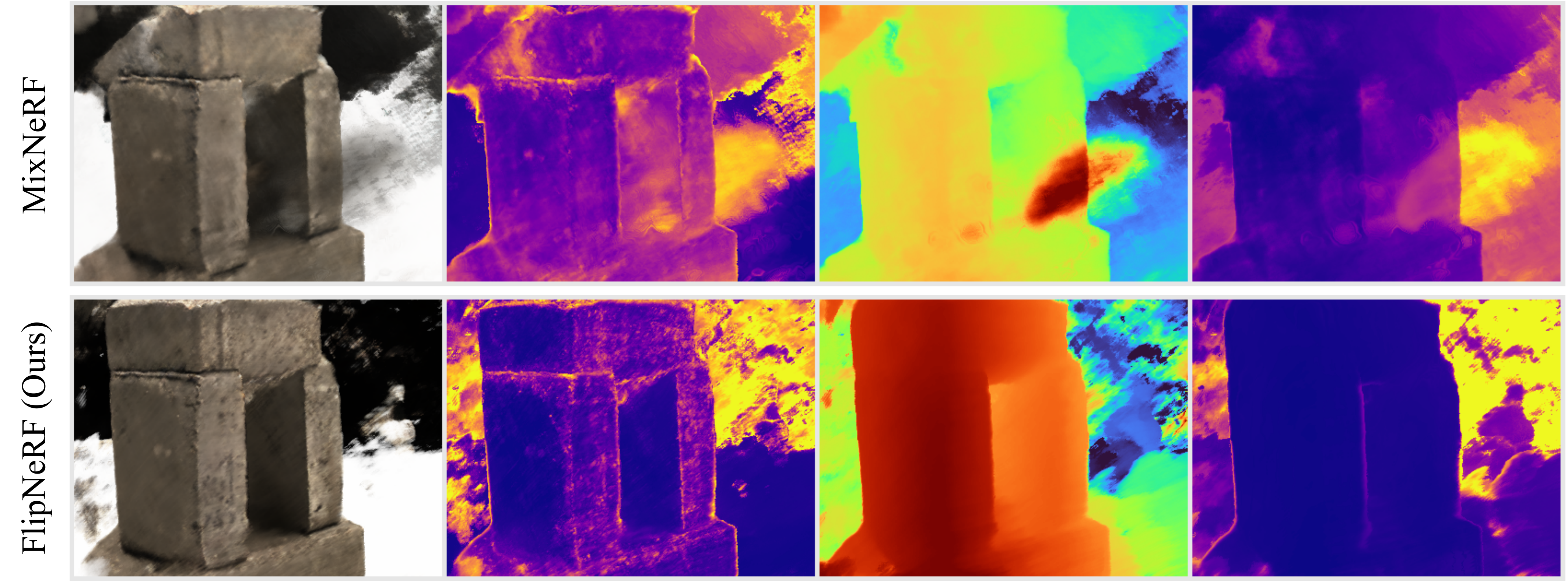}
    \vspace{-2mm}
    \begin{tabular}{p{0.00001\linewidth}p{0.18\linewidth}p{0.18\linewidth}p{0.19\linewidth}p{0.2\linewidth}}
     & \centering\scriptsize RGB Image & \centering\scriptsize RGB Std. & \centering\scriptsize Depth Map & \centering\scriptsize Depth Std.
    \end{tabular}
    \end{subfigure}
     \vspace{-4mm}
    \caption{
    \textbf{Comparison of FlipNeRF and MixNeRF.}
    Our FlipNeRF renders the images from novel views with much fewer artifacts and more accurate depth maps than MixNeRF.
    Considering the standard deviation of RGB and depths, our FlipNeRF is able to estimate more reliable outputs than MixNeRF.
    For the std. map, the darker the pixel is, the more certain the output is.
    }
    \label{fig:uncertainty}
\end{figure}

\paragraph{Uncertainty-aware regularization.}
Tab.~\ref{tab:anal_ueloss} shows the effectiveness of our proposed UE Loss.
Compared to MixNeRF~\cite{seo2023mixnerf} which models a ray with mixture of distributions as our FlipNeRF, ours consistently achieves more reliable rendering results with much lower NLL.
Without $\mathcal{L}_\text{Emp.}$~\cite{wang2022score} or our proposed $\mathcal{L}_\text{UE}$, ours already outperforms MixNeRF by a large margin.
However, ours with naively leveraged $\mathcal{L}_\text{Emp.}$ rather shows inferior results to MixNeRF.
It shows that naive application of the existing regularization technique for reducing artifacts under the few-shot setting can lead to overall performance degradation due to the scene-by-scene various structures.
As illustrated in Fig.~\ref{fig:uncertainty}, with our proposed $\mathcal{L}_\text{UE}$, ours improves both of the rendering quality and the reliability of the model outputs by a large margin compared to MixNeRF.

\begin{table}[t]
\centering
\resizebox{\linewidth}{!}{
\begin{tabular}{l|c|c|c|c}
\toprule
  & PSNR $\uparrow$ & SSIM $\uparrow$ & LPIPS $\downarrow$ & Average Err. $\downarrow$ \\
\midrule
w/o $\mathcal{L}_\text{BFC}$ & 18.79 & \cellcolor{red!25}\underline{0.755} & 0.200 & 0.115 \\
\midrule
w/ $\mathcal{L}_\text{BFC}$ & & & & \\
\quad MSE on $\mathbf{b}$ & \cellcolor{red!25}\underline{18.83} & \cellcolor{red!25}\underline{0.755} & 0.197 & \cellcolor{red!25}\underline{0.113} \\
\quad Cos. Sim. on $\mathbf{b}$ & 18.77 & \cellcolor{red!25}\underline{0.755} & \cellcolor{red!25}\underline{0.193} & 0.115 \\
\quad JSD on $\mathbf{b}_\mathbf{\hat{d}}$ & 18.43 & 0.749 & 0.204 & 0.120 \\
\quad JSD on $\mathbf{b}$ & \cellcolor{red!50}\textbf{19.55} & \cellcolor{red!50}\textbf{0.767} & \cellcolor{red!50}\textbf{0.180} & \cellcolor{red!50}\textbf{0.101} \\
\bottomrule
\end{tabular}
}
\vspace{-2mm}
\caption{
    \textbf{Comparison of different strategies for $\mathcal{L}_\text{BFC}$.}
    Our $\mathcal{L}_\text{BFC}$ achieves a significant performance gain compared to other regularization schemes for feature consistency.
    }
    \label{tab:anal_con_loss}
\vspace{-2mm}
\end{table}

\begin{table}[t]
\centering
\resizebox{\linewidth}{!}{
\begin{tabular}{c|ccccc|c|c|c|c}
\toprule
& $\mathcal{L}_\text{NLL}$ & $\mathcal{L}_\text{NLL}'$ & ${\mathcal{L}_\text{UE}}^\dagger$ & $\mathcal{L}_\text{BFC}$ & $\mathcal{L}_\text{Ori.}$ & {PSNR $\uparrow$} & {SSIM $\uparrow$} & {LPIPS $\downarrow$} & {Average $\downarrow$} \\
\midrule
(1) & \checkmark & & & & & 17.44 & 0.729 & 0.187 & 0.130 \\
\midrule
(2) & \checkmark & \checkmark & & & & 16.94 & 0.727 & 0.217 & 0.143 \\
(3) & \checkmark & \checkmark & \checkmark & & & 17.89 & 0.736 & 0.206 & 0.130 \\
(4) & \checkmark & \checkmark & \checkmark & \checkmark & & 19.01 & 0.755 & \cellcolor{red!25}\underline{0.181} & \cellcolor{red!25}\underline{0.107} \\
(5) & \checkmark & \checkmark & \checkmark & & \checkmark & 18.79 & 0.755 & 0.200 & 0.115 \\
(6) & \checkmark & \checkmark & & \checkmark & \checkmark & \cellcolor{red!25}\underline{19.30} & \cellcolor{red!25}\underline{0.758} & 0.196 & 0.108 \\
(7) & \checkmark & \checkmark & \checkmark & \checkmark & \checkmark & \cellcolor{red!50}\textbf{19.55} & \cellcolor{red!50}\textbf{0.767} & \cellcolor{red!50}\textbf{0.180} & \cellcolor{red!50}\textbf{0.101} \\
\bottomrule
\end{tabular}
}
\vspace{-2mm}
\caption{
\textbf{Ablation study.}
$\dagger$ indicate that the losses are applied to both $\mathbf{r}$ and $\mathbf{r}'$.
}
\label{tab:ablation}
\end{table}

\paragraph{Bottleneck feature consistency.}
As shown in Tab.~\ref{tab:anal_con_loss}, there is no significant impact on the performance with regularizing the bottleneck feature $\mathbf{b}$ using MSE or cosine similarity.
By our proposed $\mathcal{L}_\text{BFC}$ with JSD, we achieve a considerable performance improvement.
Interestingly, the performance rather degrades when we apply $\mathcal{L}_{\text{BFC}}$ with JSD to $\mathbf{b}_{\hat{\mathbf{d}}}$, \ie the bottleneck feature conditioned with input viewing direction $\hat{\mathbf{d}}$.
We conjecture that the reduced feature dimension of $\mathbf{b}_{\hat{\mathbf{d}}}$ reduces the capacity of feature representations and prevents the model from improving robustness.

\subsection{Ablation Study}
\label{subsec:ablation}
The quantitative results of our ablation study are reported in Tab.~\ref{tab:ablation}.
With only additionally exploiting our proposed flipped reflection rays, our FlipNeRF achieves more degenerate results compared to the baseline ((1) $\rightarrow$ (2)).
However, we are able to achieve performance improvement by a large margin with our proposed $\mathcal{L}_\text{UE}$ and $\mathcal{L}_\text{BFC}$ ((2) $\rightarrow$ (3) $\rightarrow$ (4)).
(5) $\rightarrow$ (7) shows that enhancing the consistency of bottleneck features between the pair of original and flipped reflection rays is considerably effective.
Additionally, by leveraging $\mathcal{L}_\text{Ori.}$ from Ref-NeRF, we are able to estimate more accurate normal vectors, leading to more effective flipped reflection rays and performance gain ((4) $\rightarrow$ (7)).

\begin{table*}[t]
\centering
\resizebox{0.9\linewidth}{!}{
\begin{tabular}{l|cc|cc|cc|cc|cc}
\toprule
  \multirow{2}{*}{} & \multicolumn{2}{c}{PSNR $\uparrow$} & \multicolumn{2}{c}{SSIM $\uparrow$} & \multicolumn{2}{c}{LPIPS $\downarrow$} & \multicolumn{2}{c}{Average Err. $\downarrow$} & \multicolumn{2}{c}{MAE$^\circ$ $\downarrow$}  \\
  & 4-view & 8-view & 4-view & 8-view & 4-view & 8-view & 4-view & 8-view & 4-view & 8-view \\ \midrule
mip-NeRF~\cite{barron2021mip} & 14.12 & 18.74 & 0.722 & 0.828 & 0.382 & 0.238 & 0.221 & 0.121 & 96.05 & 101.21 \\
Ref-NeRF~\cite{verbin2022ref} & 18.09 & \cellcolor{red!25}\underline{24.00} & 0.764 & \cellcolor{red!25}\underline{0.879} & 0.269 & 0.106 & 0.150 & \cellcolor{red!25}\underline{0.058} & 65.62 & \cellcolor{red!25}\underline{57.93} \\
\midrule
DietNeRF~\cite{jain2021putting} & 15.42 & 21.31 & 0.730 & 0.847 & 0.314 & 0.153 & 0.201 & 0.086 & - & - \\
InfoNeRF~\cite{kim2022infonerf} & 18.44 & 22.01 & 0.792 & 0.852 & 0.223 & 0.133 & 0.119 & 0.073 & - & - \\
RegNeRF~\cite{niemeyer2022regnerf} & 13.71 & 19.11 & 0.786 & 0.841 & 0.346 & 0.200 & 0.210 & 0.122 & \cellcolor{red!25}\underline{62.78} & 60.37 \\
MixNeRF~\cite{seo2023mixnerf} & \cellcolor{red!25}\underline{18.99} & 23.84 & \cellcolor{red!25}\underline{0.807} & 0.878 & \cellcolor{red!25}\underline{0.199} & \cellcolor{red!25}\underline{0.103} & \cellcolor{red!25}\underline{0.113} & 0.060 & 70.90 & 62.04 \\
\textbf{FlipNeRF (Ours)} & \cellcolor{red!50}\textbf{20.60} & \cellcolor{red!50}\textbf{24.38} & \cellcolor{red!50}\textbf{0.822} & \cellcolor{red!50}\textbf{0.883} & \cellcolor{red!50}\textbf{0.159} & \cellcolor{red!50}\textbf{0.095} & \cellcolor{red!50}\textbf{0.091} & \cellcolor{red!50}\textbf{0.055} & \cellcolor{red!50}\textbf{58.72} & \cellcolor{red!50}\textbf{57.17} \\
\bottomrule
\end{tabular}
}
\vspace{-2mm}
\caption{
    \textbf{Quantitative results on Realistic Synthetic 360$^\circ$.}
    Our FlipNeRF achieves the SOTA performance among other baselines across all the scenarios and metrics.
    }
    \label{tab:blender}
\vspace{-2mm}
\end{table*}

\begin{table*}[t]
\centering
\resizebox{0.95\linewidth}{!}{
\begin{tabular}{l|ccc|ccc|ccc|ccc}
\toprule
  & \multicolumn{3}{c}{PSNR $\uparrow$} & \multicolumn{3}{c}{SSIM $\uparrow$} & \multicolumn{3}{c}{LPIPS $\downarrow$} & \multicolumn{3}{c}{Average Error $\downarrow$}  \\
  & 3-view & 6-view & 9-view  & 3-view & 6-view & 9-view  & 3-view & 6-view & 9-view  & 3-view & 6-view & 9-view \\
  \midrule
 mip-NeRF~\cite{barron2021mip} & 8.68 & 16.54 & 23.58 & 0.571 & 0.741 & 0.879 & 0.353 & 0.198 & 0.092 & 0.323 & 0.148 & 0.056 \\
\midrule
\multicolumn{13}{l}{\textit{\textbf{Pre-training.}}} \\
\midrule\midrule
PixelNeRF~\cite{yu2021pixelnerf} & 16.82 & 19.11 & 20.40 & 0.695 & 0.745 & 0.768 & 0.270 & 0.232 & 0.220 & 0.147 & 0.115 & 0.100 \\
PixelNeRF$^\dagger$~\cite{yu2021pixelnerf} & \cellcolor{red!25}\underline{18.95} & 20.56 & 21.83 & 0.710 & 0.753 & 0.781 & 0.269 & 0.223 & 0.203 & 0.125 & 0.104 & 0.090 \\
SRF~\cite{chibane2021stereo} & 15.32 & 17.54 & 18.35 & 0.671 & 0.730 & 0.752 & 0.304 & 0.250 & 0.232 & 0.171 & 0.132 & 0.120 \\
SRF$^\dagger$~\cite{chibane2021stereo} & 15.68 & 18.87 & 20.75 & 0.698 & 0.757 & 0.785 & 0.281 & 0.225 & 0.205 & 0.162 & 0.114 & 0.093 \\
MVSNeRF~\cite{chen2021mvsnerf} & 18.63 & 20.70 & 22.40 & \cellcolor{red!50}\textbf{0.769} & 0.823 & 0.853 & 0.197 & 0.156 & 0.135 & 0.113 & 0.088 & 0.068 \\
MVSNeRF$^\dagger$~\cite{chen2021mvsnerf} & 18.54 & 20.49 & 22.22 & \cellcolor{red!50}\textbf{0.769} & 0.822 & 0.853 & 0.197 & 0.155 & 0.135 & 0.113 & 0.089 & 0.069 \\
\midrule
\multicolumn{13}{l}{\textit{\textbf{Regularization.}}} \\
\midrule\midrule
DietNeRF~\cite{jain2021putting} & 11.85 & 20.63 & 23.83 & 0.633 & 0.778 & 0.823 & 0.314 & 0.201 & 0.173 & 0.243 & 0.101 & 0.068 \\
RegNeRF~\cite{niemeyer2022regnerf} & 18.89 & 22.20 & 24.93 & 0.745 & \cellcolor{red!50}\textbf{0.841} & \cellcolor{red!50}\textbf{0.884} & \cellcolor{red!25}\underline{0.190} & 0.117 & 0.089 & \cellcolor{red!25}\underline{0.112} & 0.071 & 0.047 \\
MixNeRF~\cite{seo2023mixnerf} & \cellcolor{red!25}\underline{18.95} & \cellcolor{red!25}\underline{22.30} & \cellcolor{red!25}\underline{25.03} & 0.744 & 0.835 & 0.879 & 0.203 & \cellcolor{red!25}\underline{0.102} & \cellcolor{red!25}\underline{0.065} & 0.113 & \cellcolor{red!25}\underline{0.066} & \cellcolor{red!25}\underline{0.042} \\
\textbf{FlipNeRF (Ours)} & \cellcolor{red!50}\textbf{19.55} & \cellcolor{red!50}\textbf{22.45} & \cellcolor{red!50}\textbf{25.12} & \cellcolor{red!25}\underline{0.767} & \cellcolor{red!25}\underline{0.839} & \cellcolor{red!25}\underline{0.882} & \cellcolor{red!50}\textbf{0.180} & \cellcolor{red!50}\textbf{0.098} & \cellcolor{red!50}\textbf{0.062} & \cellcolor{red!50}\textbf{0.101} & \cellcolor{red!50}\textbf{0.064} & \cellcolor{red!50}\textbf{0.041} \\
\bottomrule
\end{tabular}}
\vspace{-2mm}
\caption{
    \textbf{Comparison with baselines on DTU.}
    Our FlipNeRF outperforms all the pre-training and regularization methods in every scenario, especially by a large margin under the 3-view setting.
    $\dagger$ indicates fine-tuning.
    }
    \label{tab:dtu}
\vspace{-2mm}
\end{table*}

\begin{table}[t]
\centering
\resizebox{\linewidth}{!}{
\begin{tabular}{l|c|c|c|c}
\toprule
  & PSNR $\uparrow$ & SSIM $\uparrow$ & LPIPS $\downarrow$ & Average Err. $\downarrow$ \\
  \midrule
mip-NeRF~\cite{barron2021mip} & 14.62 & 0.351 & 0.495 & 0.246 \\
\midrule
\multicolumn{5}{l}{\textit{\textbf{Pre-training.}}} \\
\midrule\midrule
PixelNeRF~\cite{yu2021pixelnerf} & 7.93 & 0.272 & 0.682 & 0.461 \\
PixelNeRF$^\dagger$~\cite{yu2021pixelnerf} & 16.17 & 0.438 & 0.512 & 0.217 \\
SRF~\cite{chibane2021stereo} & 12.34 & 0.250 & 0.591 & 0.313 \\
SRF$^\dagger$~\cite{chibane2021stereo} & 17.07 & 0.436 & 0.529 & 0.203 \\
MVSNeRF~\cite{chen2021mvsnerf} & 17.25 & 0.557 & 0.356 & 0.171 \\
MVSNeRF$^\dagger$~\cite{chen2021mvsnerf} & 17.88 & 0.584 & 0.327 & 0.157 \\
\midrule 
\multicolumn{5}{l}{\textit{\textbf{Regularization.}}} \\
\midrule\midrule
DietNeRF~\cite{jain2021putting} & 14.94 & 0.370 & 0.496 & 0.240 \\
RegNeRF~\cite{niemeyer2022regnerf} & 19.08 & 0.587 & 0.336 & 0.146 \\
MixNeRF~\cite{seo2023mixnerf} & \cellcolor{red!25}\underline{19.27} & \cellcolor{red!25}\underline{0.629} & \cellcolor{red!25}\underline{0.236} & \cellcolor{red!25}\underline{0.124} \\
\textbf{FlipNeRF (Ours)} & \cellcolor{red!50}\textbf{19.34} & \cellcolor{red!50}\textbf{0.631} & \cellcolor{red!50}\textbf{0.235} & \cellcolor{red!50}\textbf{0.123} \\
\bottomrule
\end{tabular}}
\vspace{-2mm}
\caption{
    \textbf{Comparison with baselines on LLFF 3-view.}
    $\dagger$ indicates fine-tuning.
    }
    \label{tab:llff}
\vspace{-4mm}
\end{table}

\subsection{Comparison with other SOTA Methods}
\begin{figure*}[t]
\centering
     \begin{tabular}{p{0.02\textwidth}p{0.128\textwidth}p{0.128\textwidth}p{0.128\textwidth}p{0.128\textwidth}p{0.128\textwidth}p{0.128\textwidth}p{0.03\textwidth}}
     & \centering\scriptsize mip-NeRF~\cite{barron2021mip} & \centering\scriptsize Ref-NeRF~\cite{verbin2022ref} & \centering\scriptsize  RegNeRF~\cite{niemeyer2022regnerf} & \centering\scriptsize MixNeRF~\cite{seo2023mixnerf} & \centering\scriptsize FlipNeRF (Ours) & \centering\scriptsize Ground Truth &
    \end{tabular}
\begin{subfigure}[b]{0.92\textwidth}
         \centering
        \vspace{-.07cm}
        \includegraphics[width=\linewidth]{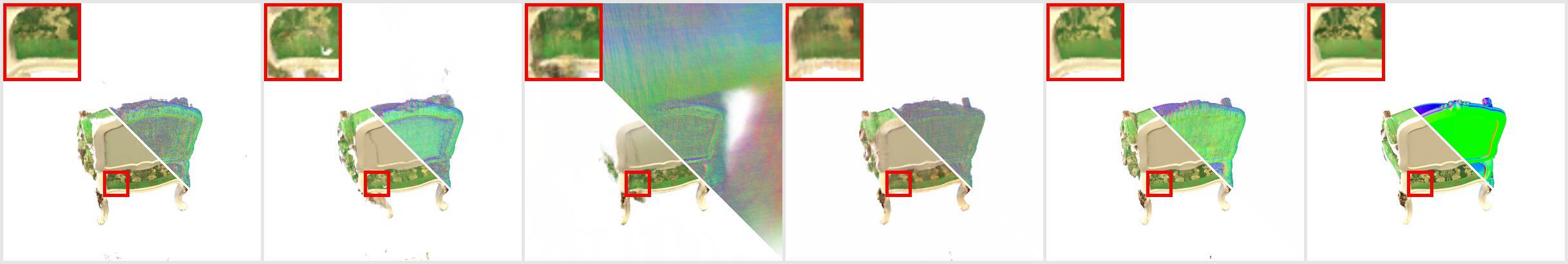}
        \vspace{-.52cm}
         \caption{Realistic Synthetic 360$^\circ$ 4-view}
         \label{fig:blender}
     \end{subfigure}
     \begin{tabular}{p{0.02\textwidth}p{0.128\textwidth}p{0.128\textwidth}p{0.128\textwidth}p{0.128\textwidth}p{0.128\textwidth}p{0.128\textwidth}p{0.03\textwidth}}
     & \centering\scriptsize mip-NeRF~\cite{barron2021mip} & \centering\scriptsize PixelNeRF~\cite{yu2021pixelnerf} & \centering\scriptsize  RegNeRF~\cite{niemeyer2022regnerf} & \centering\scriptsize MixNeRF~\cite{seo2023mixnerf} & \centering\scriptsize FlipNeRF (Ours) & \centering\scriptsize Ground Truth &
    \end{tabular}
\begin{subfigure}[b]{0.92\textwidth}
         \centering
        \vspace{-.07cm}
        \includegraphics[width=\linewidth]{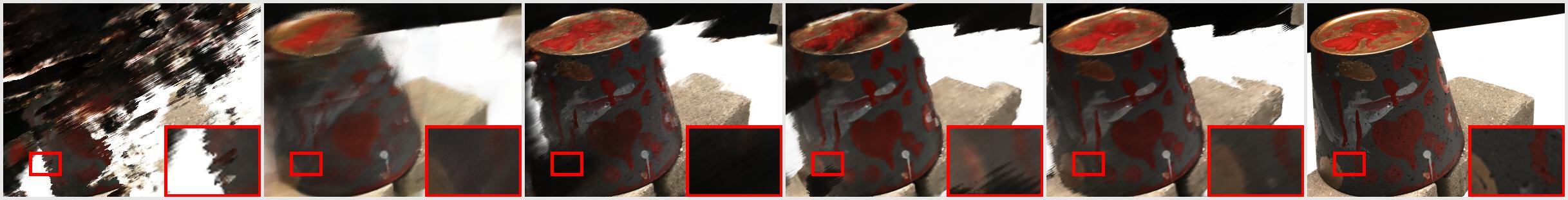}
        \vspace{-.52cm}
         \caption{DTU 3-view}
         \label{fig:dtu}
     \end{subfigure}
\vspace{-2mm}
\caption{\textbf{Qualitative results on Realistic Synthetic 360$^\circ$ and DTU.}
More results are provided in the supp. material.}
\vspace{-4mm}
\label{fig:qual_res}
\end{figure*}

\paragraph{Realistic Synthetic 360$^\circ$.}
As demonstrated in Tab.~\ref{tab:blender}, our FlipNeRF achieves the SOTA performance across all the evaluation metrics. Compared to MixNeRF which leverages a mixture model framework as ours, our FlipNeRF improves the performance by a large margin.
Noticeably, our FlipNeRF estimates more accurate surface normals than other baselines, leading to the performance gain with better reconstructed fine details from limited input views as shown in Fig.~\ref{fig:blender}.
Additionally, the vanilla Ref-NeRF, which shows great performance with accurate normal vectors, achieves comparable or even better performance than other regularization methods except ours.
From this result, we are able to expect that estimating the accurate surface normals is one of the key factors for learning 3D geometry with sparse inputs.
Note that the comparable MAE$^\circ$ of RegNeRF results from the overly-smoothed depth estimation, not indicating the high-quality of rendering results, as shown in Fig.~\ref{fig:blender}.

\paragraph{DTU.}
Our FlipNeRF achieves the best results across all the scenarios and most of the evaluation metrics on DTU as shown in Tab.~\ref{tab:dtu}.
Remarkably, ours trained with 6-view outperforms all the pre-training methods trained with 9-view in every metric except SSIM.
Furthermore, ours trained with 3-view still outperforms PixelNeRF~\cite{yu2021pixelnerf} and SRF~\cite{chibane2021stereo} trained with 6-view.
Similar to the results on Realistic Synthetic 360$^\circ$, ours outperforms other baselines by a large margin especially under the 3-view, which is the most challenging scenario, with reducing the floating artifacts successfully as shown in Fig.~\ref{fig:dtu}.
Since the flipped reflection rays are effective training resources for unseen views, the fewer the training views are provided, the more performance gain is expected.

\paragraph{LLFF.}
Table~\ref{tab:llff} compares our FlipNeRF against other baselines on LLFF, which is a real forward-facing dataset.
Although ours achieves the SOTA performance among other baselines, it is much more marginal than those on Realistic Synthetic 360$^\circ$ and DTU.
We conjecture the reason for the marginal improvement of our proposed method can be the fact that a set of flipped reflection rays, which are able to widely cover the unseen views, are not very useful for the scenes in LLFF, where a set of camera poses are much less dynamic than other datasets.
In other words, our FlipNeRF is able to not only achieve a competitive performance for the scenes consisting of a set of simple camera poses, but also render the novel views in much higher quality for more dynamically captured scenes with only a few shots.
The rendered images are provided in the supp. material.

\section{Conclusion}
\label{sec:conclusion}
In this work, we have focused on accurate surface normals, which is another key factor for the few-shot novel view synthesis.
Our proposed FlipNeRF utilizes a set of flipped reflection rays as additional training resources, which are simply derived from the estimated normal vectors and the input ray directions.
Since it does not require any heuristic factor for unseen view generation, we are able to exploit these additional training resources with much less burden.
Furthermore, with our proposed UE Loss, FlipNeRF reduces the floating artifacts consistently across the different scene structures while considering the output uncertainty, leading to more reliable outputs.
Also, our proposed BFC Loss enhances the bottleneck feature consistency between the rays cast on the photo-consistent pixels without leveraging the off-the-shelf feature extractor, leading to performance improvement under the few-shot setting.
Our FlipNeRF achieves the SOTA performance with limited input views among the other few-shot baselines and vanilla NeRF-like models.
We expect that our work is able to open another meaningful direction for the research of few-shot novel view synthesis.

\paragraph{Limitations and future work.}
Our FlipNeRF exploits the flipped reflection rays as a set of additional training rays, leading to more accurate surface normal estimation.
Although Ref-NeRF achieved promising results with the high-quality surface reconstruction, we use mip-NeRF representation instead of Ref-NeRF for a fair comparison with other methods which are based on mip-NeRF.
However, as shown in Tab.~\ref{tab:dtu} and Fig.~\ref{fig:blender}, Ref-NeRF shows promising results without any additional consideration for few-shot setting compared to the vanilla mip-NeRF.
Like our FlipNeRF, the accurate surface normal estimation leads to the competitive performance of Ref-NeRF even in the few-shot scenarios.
Therefore, a combination of our FlipNeRF training framework with the Ref-NeRF representation or further exploration with regard to the view-dependent appearance for the few-shot novel view synthesis can be interesting directions for future research.

\paragraph{Acknowledgements.}
This work was supported by NRF (2021R1A2C3006659) and IITP (2021-0-01343) both funded by Korean Government. It was also supported by Samsung Electronics (IO201223-08260-01).

{\small
\bibliographystyle{ieee_fullname}
\bibliography{egbib}
}

\setcounter{section}{0}
\setcounter{table}{0}
\setcounter{figure}{0}
\renewcommand\thesection{\Alph{section}}
\renewcommand\thetable{\Alph{table}}
\renewcommand\thefigure{\Alph{figure}}

\twocolumn[{
\maketitle
\centering
\resizebox{0.9\linewidth}{!}{
\begin{tabular}{l|c|c|c|c|c|c}
\toprule
  & \multicolumn{2}{c|}{Real. Syn. 360$^\circ$~\cite{mildenhall2021nerf}} & \multicolumn{3}{c|}{DTU~\cite{jensen2014large}} & LLFF~\cite{mildenhall2019local} \\
  & 4-view & 8-view & 3-view & 6-view & 9-view & 3-view \\
  \midrule
  Learning Rate & \multicolumn{2}{c|}{$[1\mathrm{e}{-3}, 1\mathrm{e}{-5}]$} & \multicolumn{4}{c}{$[2\mathrm{e}{-3}, 2\mathrm{e}{-5}]$} \\
  \midrule
  Warm-up Step & 512 & 1024 & 512 & \multicolumn{2}{c|}{1024} & 512 \\
  \midrule
  Delay Multiplier & \multicolumn{6}{c}{$1\mathrm{e}{-2}$} \\
  \midrule
  $\lambda_{1} (\text{for } \mathcal{L}_\text{NLL})$ & \multicolumn{6}{c}{$[4.0, 1\mathrm{e}{-3}]$} \\
  \midrule
  $\lambda_{2} (\text{for }\mathcal{L}_\text{NLL}')$ & $[4\mathrm{e}{-1}, 1\mathrm{e}{-4}]$ & $[4\mathrm{e}{-2}, 1\mathrm{e}{-5}]$ & $[4\mathrm{e}{-1}, 1\mathrm{e}{-4}]$ & $[4\mathrm{e}{-2}, 1\mathrm{e}{-5}]$ & $[4\mathrm{e}{-3}, 1\mathrm{e}{-6}]$ & $[4\mathrm{e}{-3}, 1\mathrm{e}{-6}]$ \\
  \midrule
  $\lambda_{3} (\text{for }\mathcal{L}_\text{UE})$ & $[1\mathrm{e}{-4}, 1\mathrm{e}{-1}]$ & $[1\mathrm{e}{-5}, 1\mathrm{e}{-2}]$ & $[1\mathrm{e}{-4}, 1\mathrm{e}{-1}]$ & $[1\mathrm{e}{-5}, 1\mathrm{e}{-2}]$ & $[1\mathrm{e}{-6}, 1\mathrm{e}{-3}]$ & $[1\mathrm{e}{-6}, 1\mathrm{e}{-3}]$ \\
  \midrule
  $\lambda_{4} (\text{for }\mathcal{L}_\text{UE}')$ & $1\mathrm{e}{-2}$ & $1\mathrm{e}{-3}$ & $1\mathrm{e}{-3}$ & $1\mathrm{e}{-4}$ & $1\mathrm{e}{-5}$ & $1\mathrm{e}{-5}$ \\
  \midrule
  $\lambda_{5} (\text{for }\mathcal{L}_\text{BFC})$ & $1\mathrm{e}{-1}$ & $1\mathrm{e}{-2}$ & $1\mathrm{e}{-1}$ & $1\mathrm{e}{-2}$ & $1\mathrm{e}{-3}$ & $1\mathrm{e}{-3}$ \\
  \midrule
  $\lambda_{6} (\text{for }\mathcal{L}_\text{Ori.})$ & $1\mathrm{e}{-1}$ & $1\mathrm{e}{-2}$ & $1\mathrm{e}{-1}$ & $1\mathrm{e}{-2}$ & $1\mathrm{e}{-3}$ & $1\mathrm{e}{-3}$ \\
\bottomrule
\end{tabular}}
\vspace{-2mm}
\captionof{table}{
    \textbf{Details of hyperparameters and loss balancing terms.}
    For each dataset, the more training views are provided, the smaller $\lambda$'s we set to prevent over-regularization, except $\lambda_{1}$ for $\mathcal{L}_\text{NLL}$.
    $[a, b]$ indicates the annealing from $a$ to $b$.
    }
    \label{supp_tab:training_detail}
\vspace{3mm}
}]

\section{Implementation Details}
\label{supp_sec:impl_details}
The detailed hyperparameters and our loss balancing terms by the datasets and the number of training views are provided in Tab.~\ref{supp_tab:training_detail}.

\begin{figure}[!t]
\centering
\includegraphics[width=\linewidth]{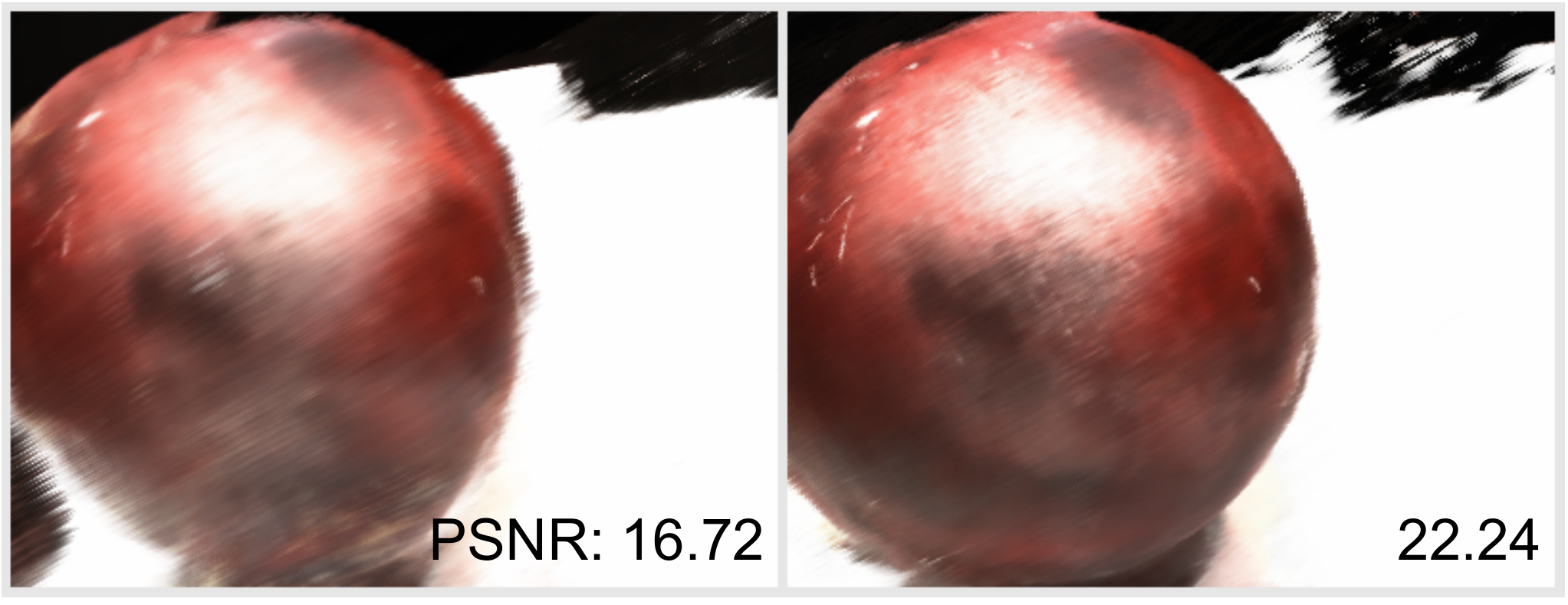}
\begin{tabular}{p{0.45\linewidth}p{0.45\linewidth}}
     \centering w/ explicit normalization & \centering w/o explicit normalization
\end{tabular}
\vspace{-5mm}
\caption{\textbf{Comparison between FlipNeRF using $\hat{\mathbf{n}}$ with and without an explicit normalization.}
With explicitly normalized $\hat{\mathbf{n}}$ (left), our FlipNeRF suffers from the training instability and achieves degenerate results.
We are able to achieve much superior rendering quality with our proposed surface normals (right), which are trained to approximate unit vectors, thanks to our regularization techniques and masking strategy.
}
\label{supp_fig:normalization_comparison}
\end{figure}

\section{Explicit Normalization for the Estimated Surface Normals}
\label{supp_sec:normalization}
\begin{figure}[!t]
\centering
\includegraphics[width=0.9\linewidth]{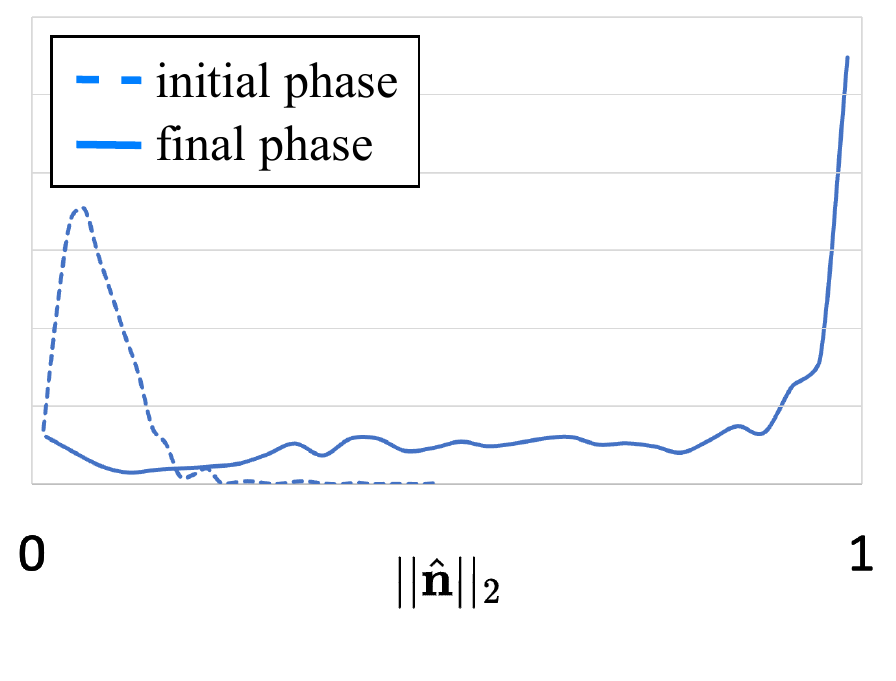}
\vspace{-5mm}
\caption{\textbf{Distribution of $\norm{\hat{\mathbf{n}}}_2$ without an explicit normalization process.}
At an initial training stage, the estimated surface normals $\hat{\mathbf{n}}$ are not unit vectors as most of $\norm{\hat{\mathbf{n}}}_2$ are far from 1.
However, our proposed $\hat{\mathbf{n}}$ is trained to be a unit vector naturally through the training without an explicit normalization process, and most of $\norm{\hat{\mathbf{n}}}_2$ are concentrated close to 1 after the training, which indicates that $\hat{\mathbf{n}}$ is successfully approximated to a unit vector.
}
\label{supp_fig:normal_length_dist}
\end{figure}

As mentioned in Sec. 3.2, we use the weighted sum of blending weights and estimated normal vectors along a ray, \ie $\hat{\mathbf{n}} = \sum_{i=1}^{M} w_{i}\mathbf{n}_{i}$, as the surface normals $\hat{\mathbf{n}}$ to derive a flipped reflection direction $\mathbf{d}'$.
In this formulation, $\hat{\mathbf{n}}$ is not guaranteed to be a unit vector without an explicit normalization process.
However, we empirically found that the normalization rather destabilizes the training and leads to the performance degradation as shown in Fig.~\ref{supp_fig:normalization_comparison}.
We conjecture that the inaccurately generated $\mathbf{r}'$ with an explicit normalization provides wrong supervisory signals, especially during the initial training stage when the surface normals are not accurately estimated.
Furthermore, as illustrated in Fig.~\ref{supp_fig:normal_length_dist}, $\hat{\mathbf{n}}$ is trained naturally to approximate the unit vector through the training without an explicit normalization process.
With our proposed loss terms and regularization techniques for accurate normal estimation, we are able to use $\hat{\mathbf{n}}$ as the surface normal vector without any additional normalization process.

\section{Additional Qualitative Results}
\label{supp_sec:additional_qual_res}
\begin{figure*}[!t]
\centering
     \begin{tabular}{p{0.02\textwidth}p{0.128\textwidth}p{0.128\textwidth}p{0.128\textwidth}p{0.128\textwidth}p{0.128\textwidth}p{0.128\textwidth}p{0.03\textwidth}}
     & \centering\scriptsize mip-NeRF~\cite{barron2021mip} & \centering\scriptsize Ref-NeRF~\cite{verbin2022ref} & \centering\scriptsize  RegNeRF~\cite{niemeyer2022regnerf} & \centering\scriptsize MixNeRF~\cite{seo2023mixnerf} & \centering\scriptsize FlipNeRF (Ours) & \centering\scriptsize Ground Truth &
    \end{tabular}
\begin{subfigure}[b]{0.92\textwidth}
         \centering
        \vspace{-.07cm}
        \includegraphics[width=\linewidth]{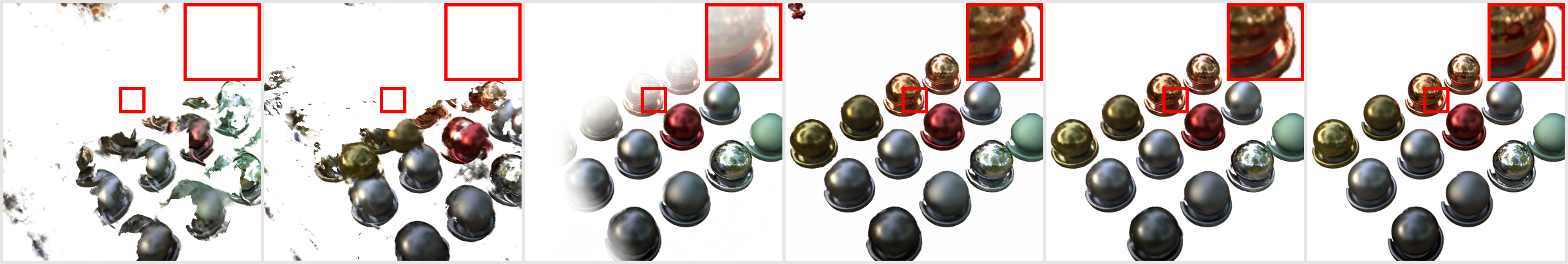}
        \vspace{-.52cm}
         \caption{4-view}
         \label{supp_fig:qual_comp_blender_4view_rgb}
     \end{subfigure}
\begin{subfigure}[b]{0.92\textwidth}
         \centering
        \includegraphics[width=\linewidth]{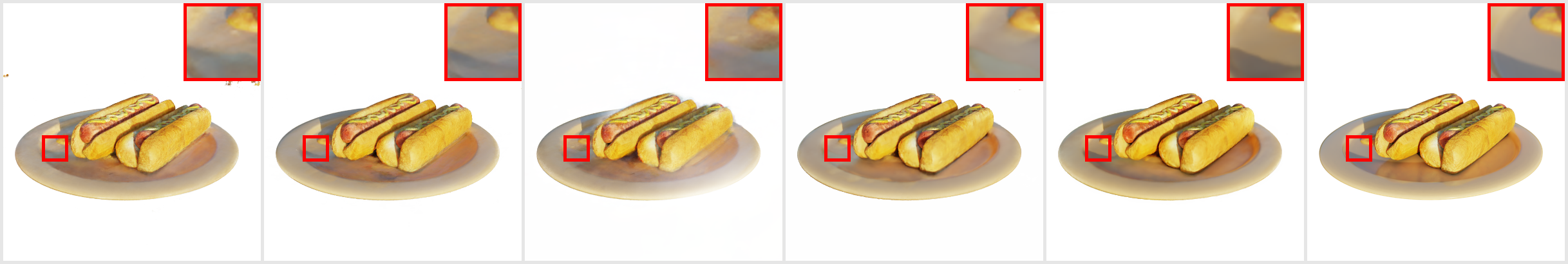}
        \vspace{-.52cm}
         \caption{8-view}
         \label{supp_fig:qual_comp_blender_8view_rgb}
     \end{subfigure}
\vspace{-2mm}
\caption{\textbf{Additional qualitative comparisons on Realistic Synthetic 360$^\circ$.}}
\label{supp_fig:qual_comp_blender}
\end{figure*}

\begin{figure*}[!t]
\centering
     \begin{tabular}{p{0.02\textwidth}p{0.128\textwidth}p{0.128\textwidth}p{0.128\textwidth}p{0.128\textwidth}p{0.128\textwidth}p{0.128\textwidth}p{0.03\textwidth}}
     & \centering\scriptsize mip-NeRF~\cite{barron2021mip} & \centering\scriptsize PixelNeRF~\cite{yu2021pixelnerf} & \centering\scriptsize  RegNeRF~\cite{niemeyer2022regnerf} & \centering\scriptsize MixNeRF~\cite{seo2023mixnerf} & \centering\scriptsize FlipNeRF (Ours) & \centering\scriptsize Ground Truth &
    \end{tabular}
\begin{subfigure}[b]{0.92\textwidth}
         \centering
        \vspace{-.07cm}
        \includegraphics[width=\linewidth]{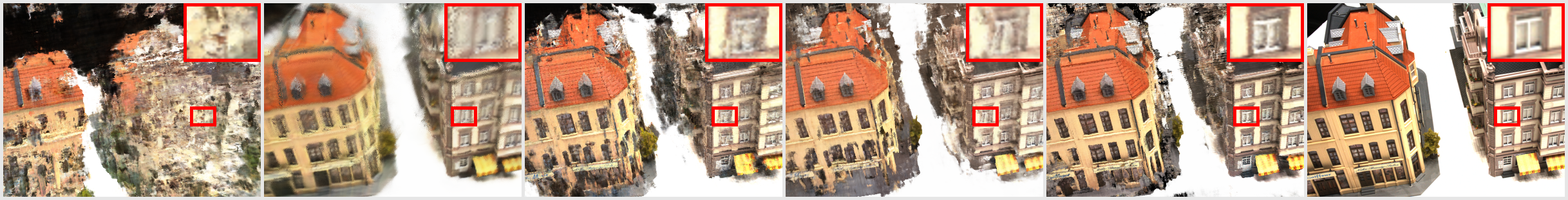}
        \vspace{-.52cm}
         \caption{3-view}
         \label{supp_fig:qual_comp_dtu_3view_rgb}
     \end{subfigure}
\begin{subfigure}[b]{0.92\textwidth}
         \centering
        \includegraphics[width=\linewidth]{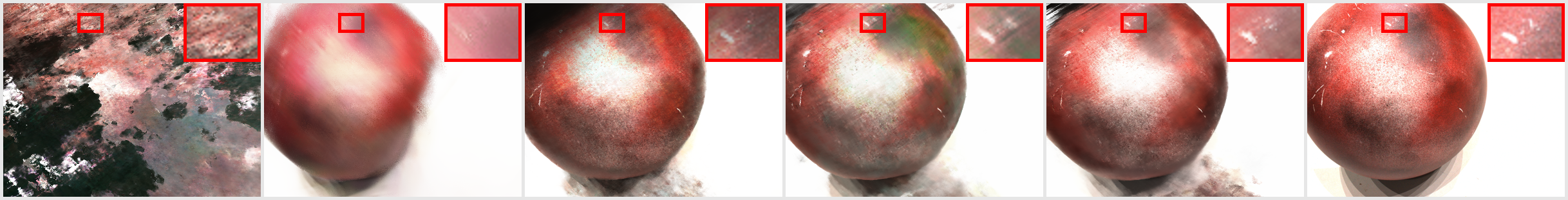}
        \vspace{-.52cm}
         \caption{6-view}
         \label{supp_fig:qual_comp_dtu_6view_rgb}
     \end{subfigure}
\begin{subfigure}[b]{0.92\textwidth}
         \centering
        \includegraphics[width=\linewidth]{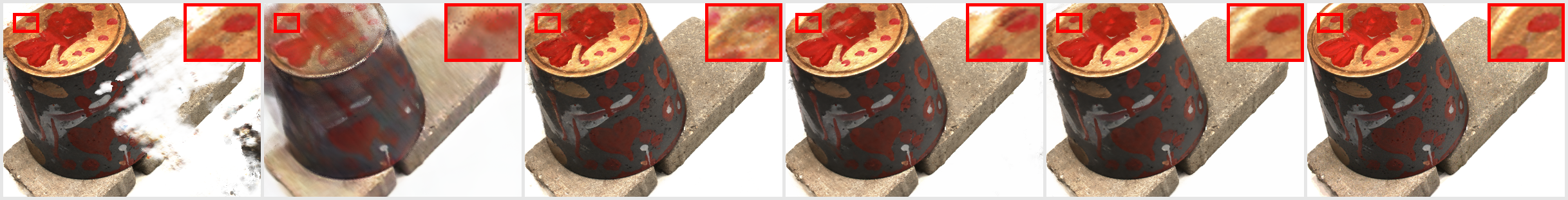}
        \vspace{-.52cm}
         \caption{9-view}
         \label{supp_fig:qual_comp_dtu_9view_rgb}
     \end{subfigure}
\vspace{-2mm}
\caption{\textbf{Additional qualitative comparisons on DTU.}}
\label{supp_fig:qual_comp_dtu}
\end{figure*}

\begin{figure*}[!t]
\centering
     \begin{tabular}{p{0.02\textwidth}p{0.128\textwidth}p{0.128\textwidth}p{0.128\textwidth}p{0.128\textwidth}p{0.128\textwidth}p{0.128\textwidth}p{0.03\textwidth}}
     & \centering\scriptsize mip-NeRF~\cite{barron2021mip} & \centering\scriptsize DietNeRF~\cite{jain2021putting} & \centering\scriptsize  RegNeRF~\cite{niemeyer2022regnerf} & \centering\scriptsize MixNeRF~\cite{seo2023mixnerf} & \centering\scriptsize FlipNeRF (Ours) & \centering\scriptsize Ground Truth &
    \end{tabular}
\begin{subfigure}[b]{0.92\textwidth}
         \centering
        \vspace{-.07cm}
        \includegraphics[width=\linewidth]{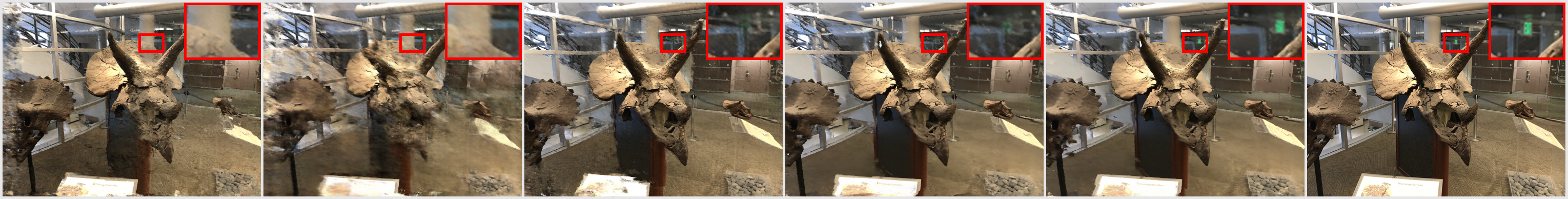}
        \vspace{-.52cm}
         \label{supp_fig:qual_comp_llff_3view_rgb}
     \end{subfigure}
\caption{\textbf{Qualitative comparisons on LLFF 3-view.}}
\label{supp_fig:qual_comp_llff}
\end{figure*}

\begin{figure*}[!t]
\centering
\begin{subfigure}[b]{0.92\textwidth}
         \centering
        \includegraphics[width=\linewidth]{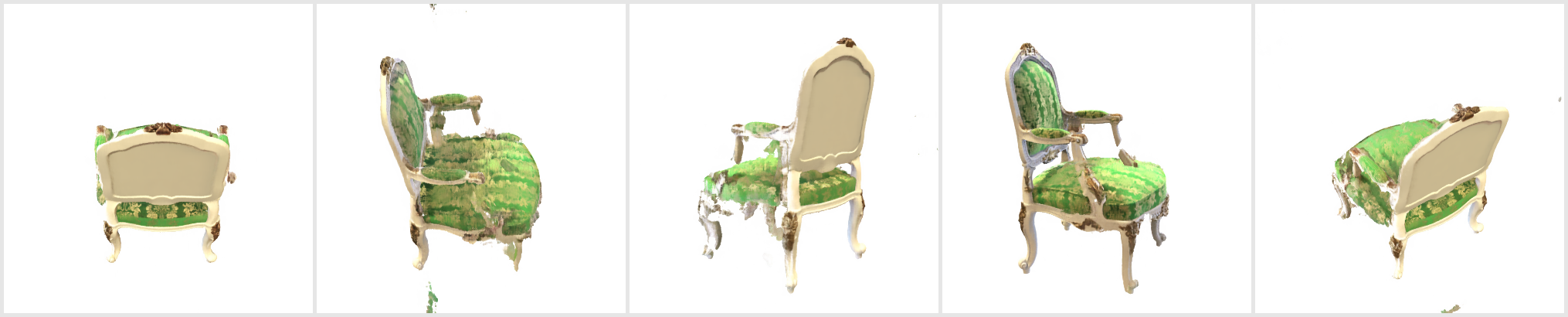}
        \vspace{-.4cm}
         \label{supp_fig:qual_res_blender_4view_1}
     \end{subfigure}
\begin{subfigure}[b]{0.92\textwidth}
         \centering
        \includegraphics[width=\linewidth]{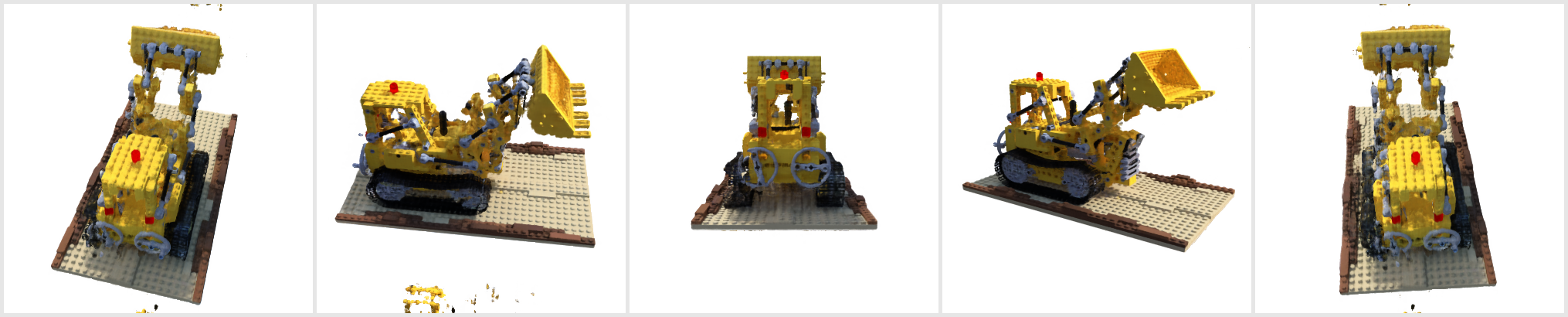}
        \vspace{-.4cm}
         \label{supp_fig:qual_res_blender_4view_2}
     \end{subfigure}
\begin{subfigure}[b]{0.92\textwidth}
         \centering
        \includegraphics[width=\linewidth]{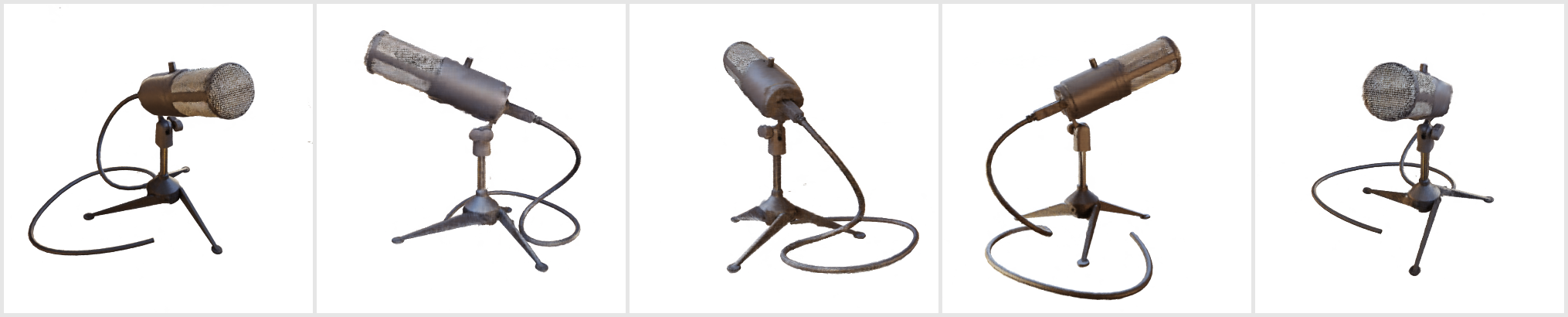}
        \vspace{-.5cm}
         \caption{4-view}
         \label{supp_fig:qual_res_blender_4view_3}
     \end{subfigure}
\begin{subfigure}[b]{0.92\textwidth}
         \centering
        \includegraphics[width=\linewidth]{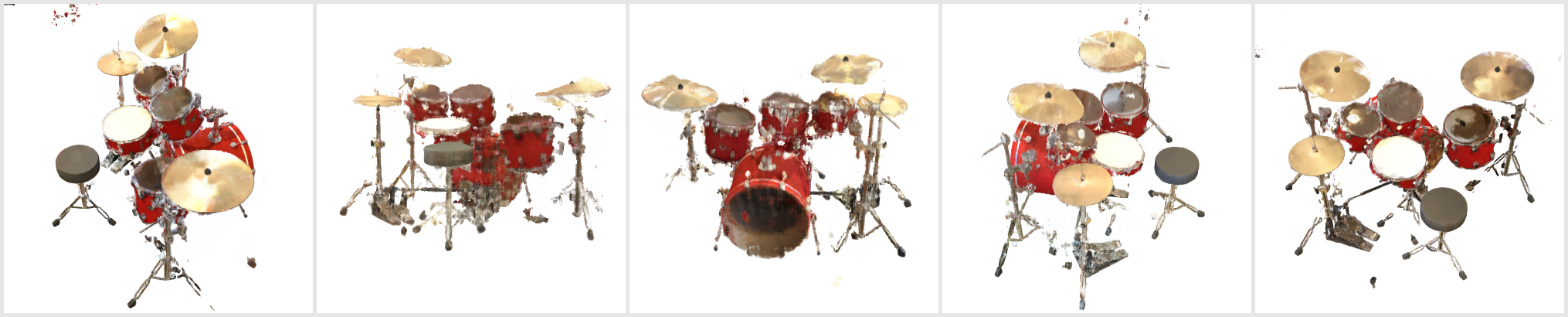}
        \vspace{-.4cm}
         \label{supp_fig:qual_res_blender_8view_1}
     \end{subfigure}
\begin{subfigure}[b]{0.92\textwidth}
         \centering
        \includegraphics[width=\linewidth]{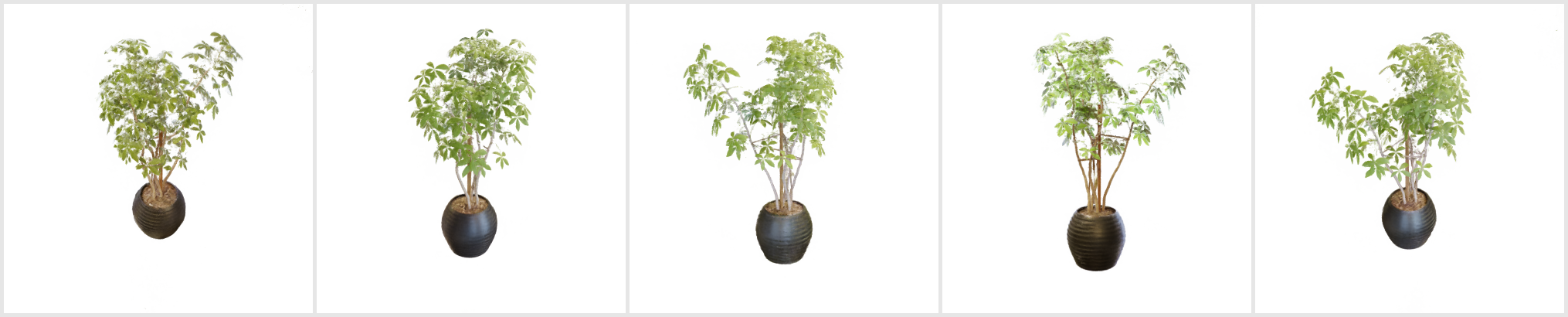}
        \vspace{-.4cm}
         \label{supp_fig:qual_res_blender_8view_2}
     \end{subfigure}
\begin{subfigure}[b]{0.92\textwidth}
         \centering
        \includegraphics[width=\linewidth]{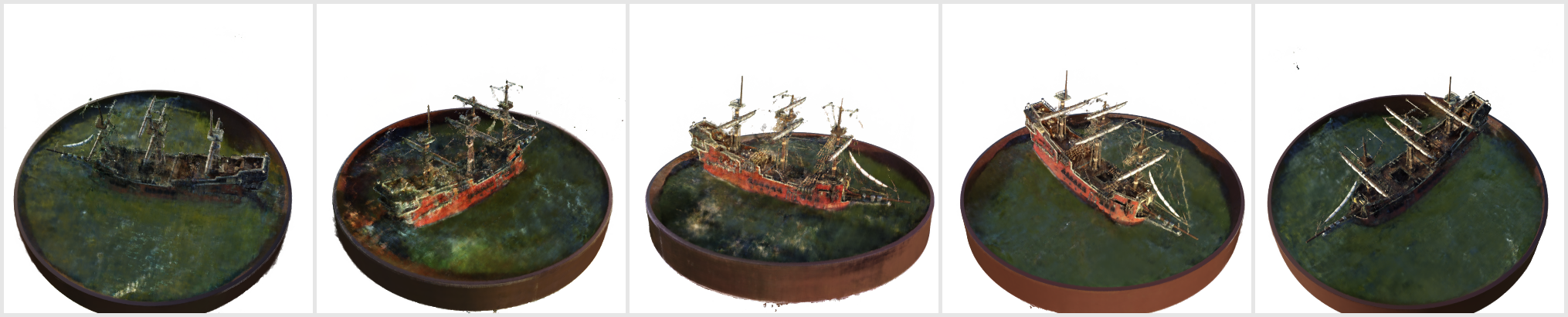}
        \vspace{-.5cm}
         \caption{8-view}
         \label{supp_fig:qual_res_blender_8view_3}
     \end{subfigure}
\vspace{-2mm}
\caption{\textbf{Additional qualitative results of our FlipNeRF on Realistic Synthetic 360$^\circ$.}}
\vspace{-4mm}
\label{supp_fig:qual_ours_blender}
\end{figure*}

\begin{figure*}[!t]
\centering
\begin{subfigure}[b]{0.8\textwidth}
         \centering
        \includegraphics[width=\linewidth]{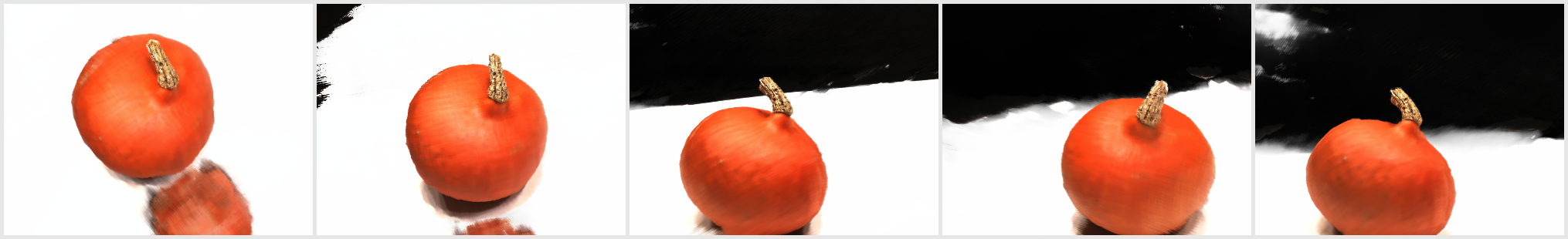}
        \vspace{-.4cm}
         \label{supp_fig:qual_res_dtu_3view_1}
     \end{subfigure}
\begin{subfigure}[b]{0.8\textwidth}
         \centering
        \includegraphics[width=\linewidth]{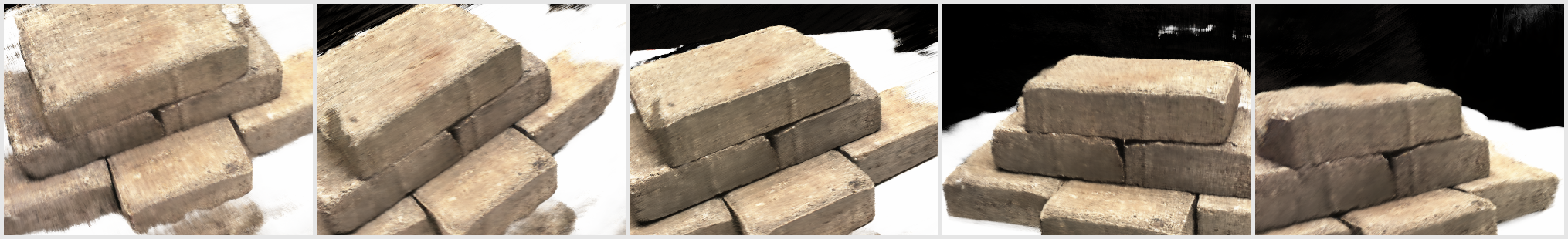}
        \vspace{-.4cm}
         \label{supp_fig:qual_res_dtu_3view_2}
     \end{subfigure}
\begin{subfigure}[b]{0.8\textwidth}
         \centering
        \includegraphics[width=\linewidth]{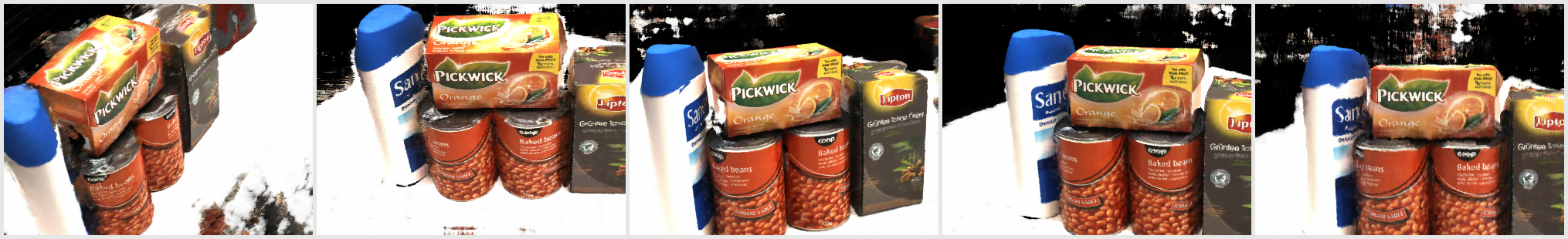}
        \vspace{-.5cm}
         \caption{3-view}
         \label{supp_fig:qual_res_dtu_3view_3}
     \end{subfigure}
\begin{subfigure}[b]{0.8\textwidth}
         \centering
        \includegraphics[width=\linewidth]{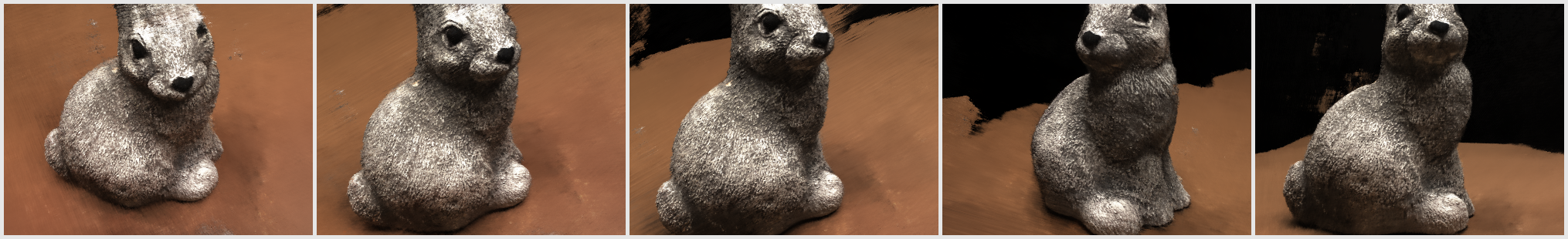}
        \vspace{-.4cm}
         \label{supp_fig:qual_res_dtu_6view_1}
     \end{subfigure}
\begin{subfigure}[b]{0.8\textwidth}
         \centering
        \includegraphics[width=\linewidth]{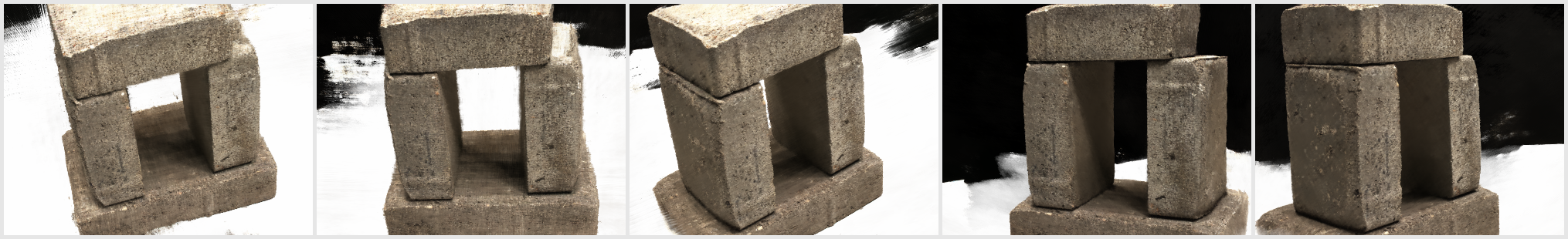}
        \vspace{-.4cm}
         \label{supp_fig:qual_res_dtu_6view_2}
     \end{subfigure}
\begin{subfigure}[b]{0.8\textwidth}
         \centering
        \includegraphics[width=\linewidth]{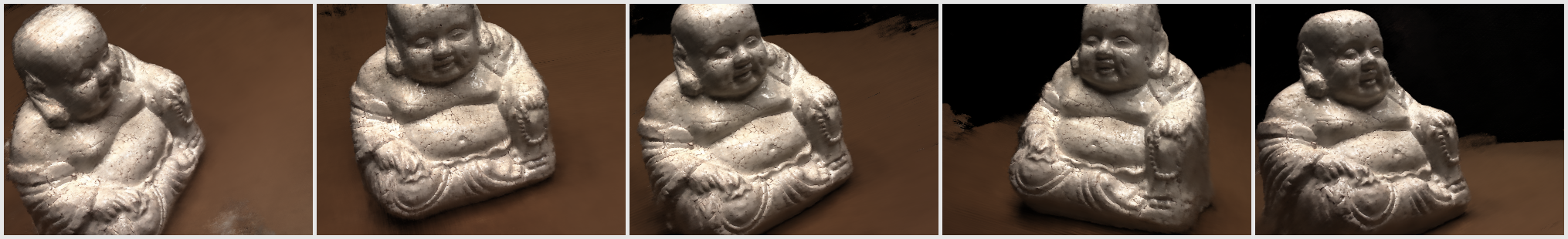}
        \vspace{-.5cm}
         \caption{6-view}
         \label{supp_fig:qual_res_dtu_6view_3}
     \end{subfigure}
\begin{subfigure}[b]{0.8\textwidth}
         \centering
        \includegraphics[width=\linewidth]{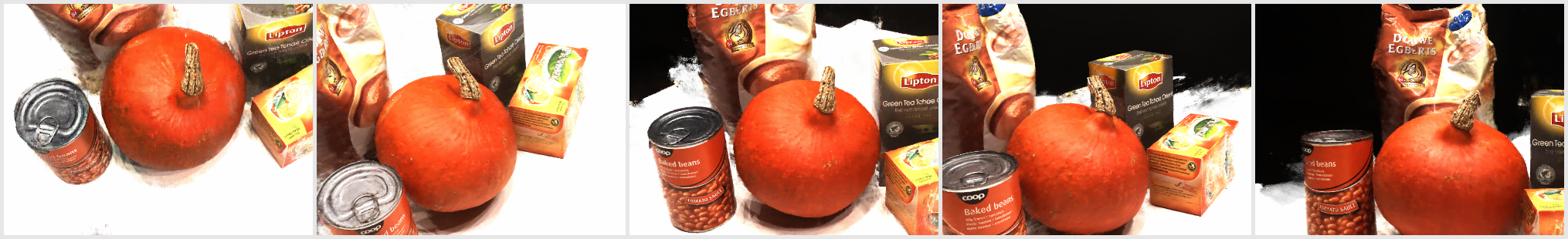}
        \vspace{-.4cm}
         \label{supp_fig:qual_res_dtu_9view_1}
     \end{subfigure}
\begin{subfigure}[b]{0.8\textwidth}
         \centering
        \includegraphics[width=\linewidth]{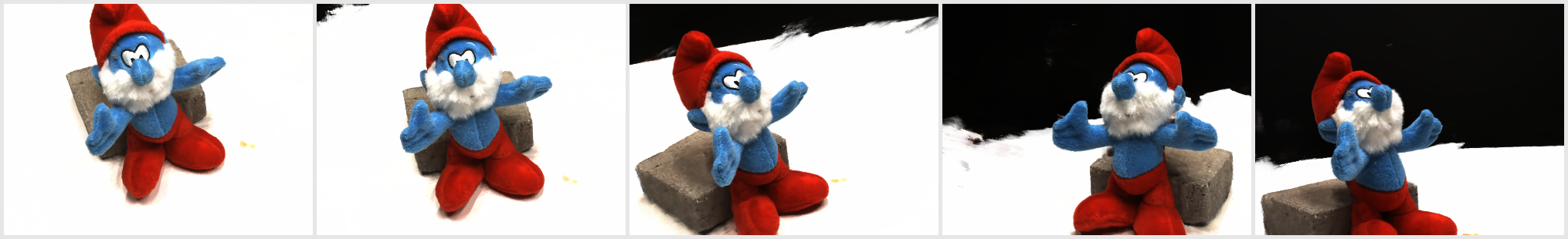}
        \vspace{-.4cm}
         \label{supp_fig:qual_res_dtu_9view_2}
     \end{subfigure}
\begin{subfigure}[b]{0.8\textwidth}
         \centering
        \includegraphics[width=\linewidth]{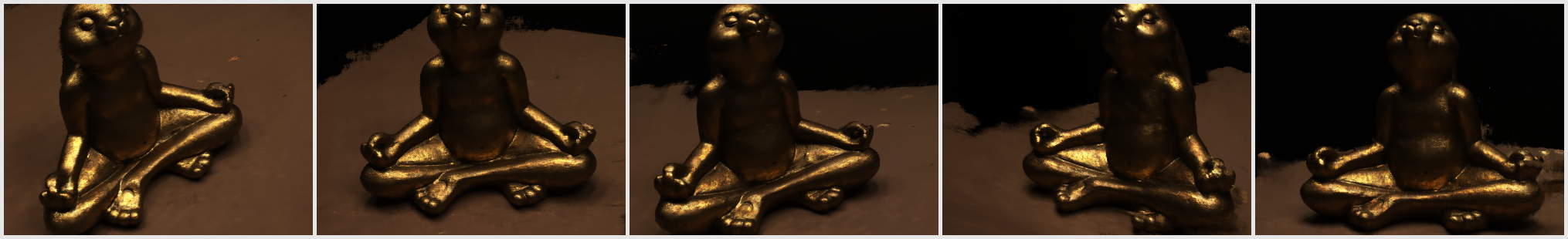}
        \vspace{-.5cm}
         \caption{9-view}
         \label{supp_fig:qual_res_dtu_9view_3}
     \end{subfigure}
\vspace{-2mm}
\caption{\textbf{Additional qualitative results of our FlipNeRF on DTU.}}
\vspace{-4mm}
\label{supp_fig:qual_ours_dtu}
\end{figure*}

\begin{figure*}[!t]
\centering
\begin{subfigure}[b]{0.8\textwidth}
         \centering
        \includegraphics[width=\linewidth]{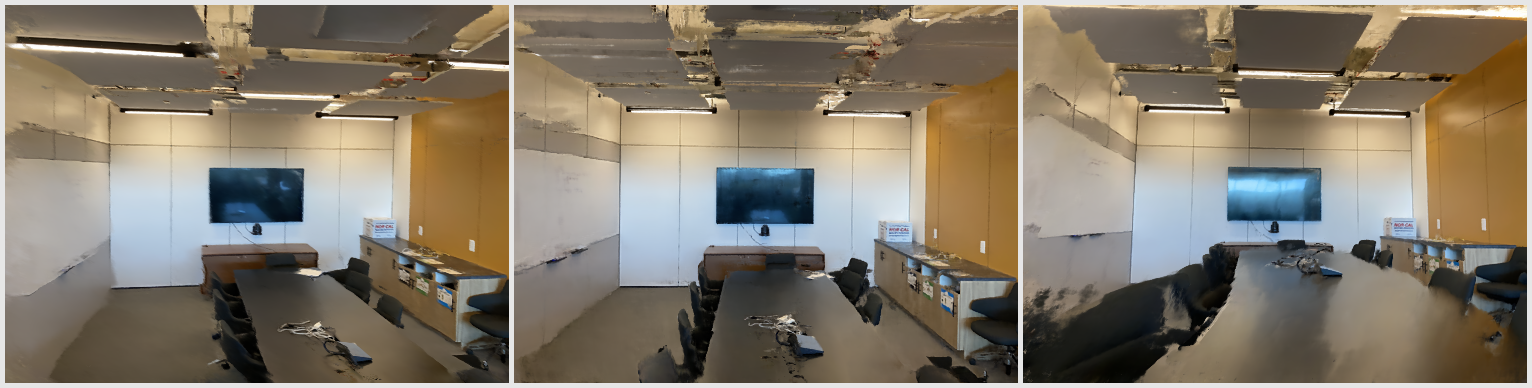}
        \vspace{-.4cm}
         \label{supp_fig:qual_res_llff_3view_1}
     \end{subfigure}
\begin{subfigure}[b]{0.8\textwidth}
         \centering
        \includegraphics[width=\linewidth]{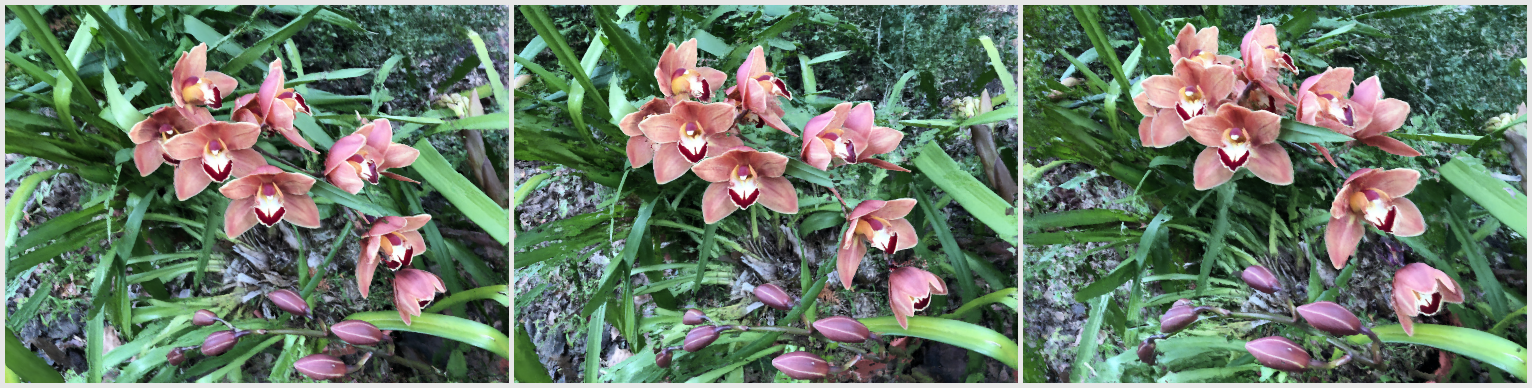}
        \vspace{-.4cm}
         \label{supp_fig:qual_res_llff_3view_2}
     \end{subfigure}
\begin{subfigure}[b]{0.8\textwidth}
         \centering
        \includegraphics[width=\linewidth]{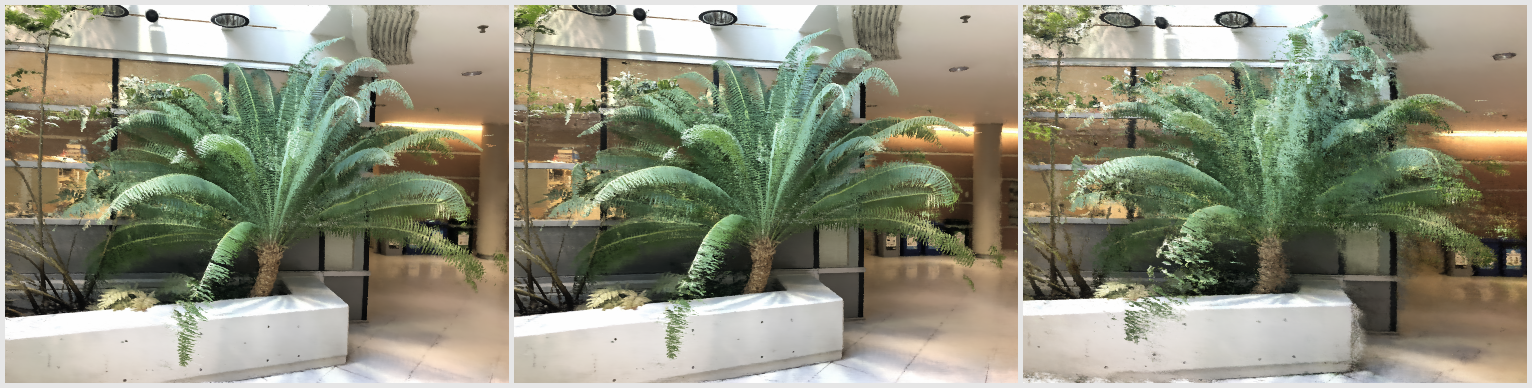}
        \vspace{-.5cm}
         \label{supp_fig:qual_res_llff_3view_3}
     \end{subfigure}
\caption{\textbf{Additional qualitative results of our FlipNeRF on LLFF 3-view.}}
\vspace{-4mm}
\label{supp_fig:qual_ours_llff}
\end{figure*}

The additional qualitative comparisons are provided in Fig.~\ref{supp_fig:qual_comp_blender}, Fig.~\ref{supp_fig:qual_comp_dtu}, and Fig.~\ref{supp_fig:qual_comp_llff}.
Furthermore, we provide more qualitative results of our FlipNeRF in Fig.~\ref{supp_fig:qual_ours_blender}, Fig.~\ref{supp_fig:qual_ours_dtu}, and Fig.~\ref{supp_fig:qual_ours_llff}.
Our FlipNeRF shows superior rendering quality compared to other baselines.

\end{document}